\pdfoutput=1

\documentclass[11pt]{article}

\usepackage[preprint]{acl}

\usepackage{times}
\usepackage{latexsym}

\usepackage[T1]{fontenc}

\usepackage[utf8]{inputenc}

\usepackage{microtype}

\usepackage{inconsolata}

\usepackage{graphicx}

\usepackage[utf8]{inputenc} %
\usepackage[T1]{fontenc}    %
\usepackage{hyperref}       %
\usepackage{url}            %
\usepackage{booktabs}       %
\usepackage{amsfonts}       %
\usepackage{nicefrac}       %
\usepackage{microtype}      %
\usepackage{xcolor}         %
\usepackage{graphicx}
\usepackage{wrapfig}
\usepackage{subcaption}
\usepackage{xspace}
\usepackage[capitalise]{cleveref}
\usepackage{enumitem}
\usepackage{tcolorbox}
\usepackage{verbatim}
\usepackage{multirow}
\usepackage{listings}

\def\eg{{\em e.g.,}\xspace}
\def\ie{{\em i.e.,}\xspace}
\def\etc{{\em etc.}\xspace}
\newcommand{\dataset}{{MMLU-Redux}\xspace}

\newcommand{\adjem}[4]{%
  \ifthenelse{\equal{#2}{#4}}{%
    #1 \scriptsize{(#2)} $\rightarrow$ \normalsize{#3} \scriptsize{(#4)}%
  }{%
    \ifthenelse{\numexpr #2 < #4}{%
      \textcolor{red}{#1 \scriptsize{(#2)} $\rightarrow$ \normalsize{#3} \scriptsize{(#4)}}%
    }{%
      \textcolor{teal}{#1 \scriptsize{(#2)} $\rightarrow$ \normalsize{#3} \scriptsize{(#4)}}%
    }%
  }%
}

\newcommand{\answerYes}[1][]{\textcolor{blue}{[Yes] #1}}
\newcommand{\answerNo}[1][]{\textcolor{orange}{[No] #1}}
\newcommand{\answerNA}[1][]{\textcolor{gray}{[N/A] #1}}
\newcommand{\answerTODO}[1][]{\textcolor{red}{\bf [TODO]}}

\newcommand{\jean}[1]{\textcolor{red}{Jean: #1}}
\newcommand{\aryo}[1]{\textcolor{orange}{Aryo: #1}}
\newcommand{\gw}[1]{\textbf{\textcolor{blue}{{gw: #1}}}}
\newcommand{\EK}[1]{\textcolor{red!69!green}{\textbf{EvK: #1}}}
\newcommand{\alb}[1]{\textcolor{teal}{\textbf{Alb: #1}}}
\newcommand{\ale}[1]{\textbf{\textcolor{olive}{{ale: #1}}}}
\newcommand{\yuzhao}[1]{\textbf{\textcolor{cyan}{{[YZ]: #1}}}}
\newcommand{\josh}[1]{\textbf{\textcolor{purple}{{JoshO: #1}}}}
\newcommand{\joshH}[1]{\textbf{\textcolor{brown}{{JoshH: #1}}}}
\newcommand{\xuanli}[1]{\textbf{\textcolor{teal}{{XH: \small{#1}}}}}

\renewcommand{\jean}[1]{}
\renewcommand{\aryo}[1]{}
\renewcommand{\gw}[1]{}
\renewcommand{\EK}[1]{}
\renewcommand{\alb}[1]{}
\renewcommand{\ale}[1]{}
\renewcommand{\yuzhao}[1]{}
\renewcommand{\josh}[1]{}
\renewcommand{\joshH}[1]{}
\renewcommand{\xuanli}[1]{}

\title{Are We Done with MMLU?}

\author{
    \textbf{Aryo Pradipta Gema}\textsuperscript{1} \;
    \textbf{Joshua Ong Jun Leang}\textsuperscript{1} \;
    \textbf{Giwon Hong}\textsuperscript{1} \;
    \textbf{Alessio Devoto}\textsuperscript{2} \\
    \textbf{Alberto Carlo Maria Mancino}\textsuperscript{2,3} \;
    \textbf{Rohit Saxena}\textsuperscript{1} \;
    \textbf{Xuanli He}\textsuperscript{4} \;
    \textbf{Yu Zhao}\textsuperscript{1} \;
    \textbf{Xiaotang Du}\textsuperscript{1} \\
    \textbf{Mohammad Reza Ghasemi Madani}\textsuperscript{5} \;
    \textbf{Claire Barale}\textsuperscript{1} \;
    \textbf{Robert McHardy}\textsuperscript{6} \;
    \textbf{Joshua Harris}\textsuperscript{7} \\
    \textbf{Jean Kaddour}\textsuperscript{4} \;
    \textbf{Emile van Krieken}\textsuperscript{1} \;
    \textbf{Pasquale Minervini}\textsuperscript{1,8}\\
    \textsuperscript{1}University of Edinburgh \quad
    \textsuperscript{2}Sapienza University of Rome \quad
    \textsuperscript{3}Polytechnic University of Bari \\
    \textsuperscript{4}University College London \quad \textsuperscript{5}University of Trento \quad
    \textsuperscript{6}AssemblyAI \\
    \textsuperscript{7}UK Health Security Agency \quad
    \textsuperscript{8}Miniml.AI \\
    \texttt{\{first.last, jong2, p.minervini\}@ed.ac.uk} \\
    \texttt{alessio.devoto@uniroma1.it} \quad \texttt{alberto.mancino@poliba.it} \\ \texttt{mr.ghasemimadani@unitn.it} \quad \texttt{joshua.harris@ukhsa.gov.uk} \\
    \texttt{\{xuanli.he, jean.kaddour.20, robert.mchardy.20\}@ucl.ac.uk}
}

\begin{document}

\maketitle

\begin{abstract}
Maybe not.
We identify and analyse errors in the popular Massive Multitask Language Understanding (MMLU) benchmark.
Even though MMLU is widely adopted, our analysis demonstrates numerous ground truth errors that obscure the true capabilities of LLMs. 
For example, we find that 57\% of the analysed questions in the \emph{Virology} subset contain errors.
To address this issue, we introduce a comprehensive framework for identifying dataset errors using a novel error annotation protocol. 
Then, we create \dataset, which is a subset of 5,700 manually re-annotated questions across all 57 MMLU subjects. 
We estimate that 6.49\% of MMLU questions contain errors.
Using \dataset, we demonstrate significant discrepancies with the model performance metrics that were originally reported.
Our results strongly advocate for revising MMLU’s error-ridden questions to enhance its future utility and reliability as a benchmark.
\url{https://huggingface.co/datasets/edinburgh-dawg/mmlu-redux-2.0}. 
\end{abstract}

\section{Introduction}
\label{sec:introduction}

The advent of transformer-based Large Language Models (LLMs) \citep{DBLP:journals/corr/abs-2303-08774,DBLP:journals/corr/abs-2312-11805,claude2,DBLP:journals/corr/abs-2305-10403,DBLP:journals/corr/abs-2307-09288, anthropic2024claude, DBLP:journals/corr/abs-2307-10169, llama3modelcard} marked a significant advancement in generative models, enabling interaction with computing devices through natural language.
This advancement rendered many earlier benchmarks and leaderboards obsolete \citep{DBLP:conf/acl/LaskarBRBJH23,DBLP:journals/corr/abs-2305-13091}, leading to the compilation of more challenging and comprehensive tests.
Among these benchmarks, Massive Multitask Language Understanding (MMLU) \citep{DBLP:conf/iclr/HendrycksBBZMSS21} has gained significant popularity: It assesses both the breadth and depth of language understanding capabilities of current LLMs across a diverse range of subjects, including mathematics, history, computer science, logic, law, \etc

\begin{figure}
  \begin{center}
    \includegraphics[width=0.5\textwidth]{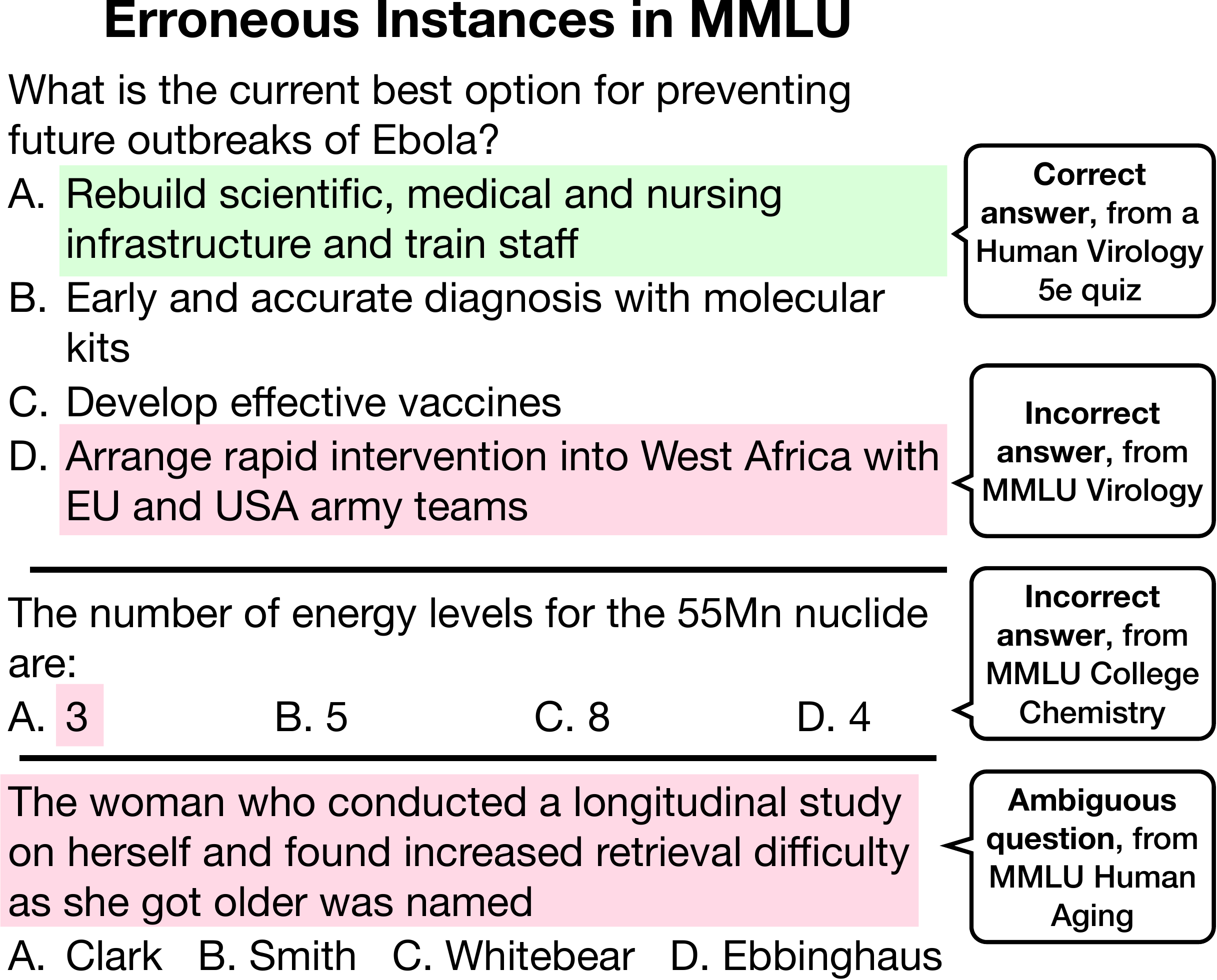}
  \end{center}
  \caption{Examples of erroneous instances from MMLU Virology, College Chemistry, and Human Aging.} 
\label{fig:err-mmlu-question}
\end{figure}

However, the reliability of benchmarking results is only as robust as the quality of the dataset used.
We find that, despite its popularity, MMLU suffers from numerous errors that can mislead evaluation and model comparison \citep{aiexplained, medium}.
These errors, which range from simple parsing and scraping mistakes to more complex issues related to context, interpretation, and dataset quality, compromise the reliability of MMLU as a benchmark. 
For example, we find that 57\% of the analysed instances in the \emph{Virology} subset contain errors, including the suggestion to send the American army to West Africa to prevent outbreaks of Ebola (see \cref{fig:err-mmlu-question}).

Therefore, in this study, we manually analyse the MMLU dataset using a novel error annotation protocol to construct \textbf{\dataset}: 14 human experts manually assessed and re-annotated 5,700 questions across all subsets of MMLU. 
We estimate that 6.49\% of the questions are erroneous. 

After our manual re-annotation effort, we study how the errors in MMLU impact LLM evaluation. 
First, we re-evaluated leading LLMs on \dataset, and found the performance metrics notably altered, changing their ranking.
Furthermore, we both quantitatively and qualitatively analysed the errors to help understand how these errors impact LLM evaluation. 
\dataset can also be used as a strong benchmark for automatic error detection in NLP datasets, which would help scale up the review of benchmark datasets. 
Therefore, we also study whether LLMs can help with error detection, using prompting techniques (\ie In-Context Learning~\citep{DBLP:journals/corr/abs-2005-14165}, Chain of Thoughts (CoT)~\citep{wei2022chain}), Retrieval Augmented Generation (RAG)~\citep{lewis2020retrieval}, and fine-tuning.
We believe \dataset underscores the need for closely studying and reassessing the benchmarks used for evaluating NLP models.

\section{What Is Wrong with MMLU?}
The MMLU dataset has become a popular choice for evaluating the performance of NLP systems, owing to its extensive coverage of subjects collected from freely available online sources with the help of graduate and undergraduate students~\citep{DBLP:conf/iclr/HendrycksBBZMSS21}.
Despite the manual effort, MMLU still contains errors that are difficult to trace due to its under-documented annotation procedure.
These errors have not yet been systematically studied, even though they were recently highlighted \citep{aiexplained,medium}.

We identify various errors in MMLU questions, from simple parsing mistakes (\eg the source answer is B, but MMLU labels it as C) to more complex issues, such as missing context.
These errors occur randomly, and without detailed documentation, tracing their root causes is challenging.

This randomness and lack of traceability highlight the need for a standardised error categorisation to improve the reliability and accuracy of the MMLU dataset.
Our approach involved developing a hierarchical annotation protocol of errors, which we used to develop \dataset: A manual annotation of all 57 subsets of MMLU, each containing 100 randomly selected samples (\cref{sec: MMLU-Redux}).

\subsection{Error Detection Annotation Protocol}

\label{sec:error_detection_taxonomy}
\begin{figure*}[t]
    \centering
    \includegraphics[width=\linewidth]{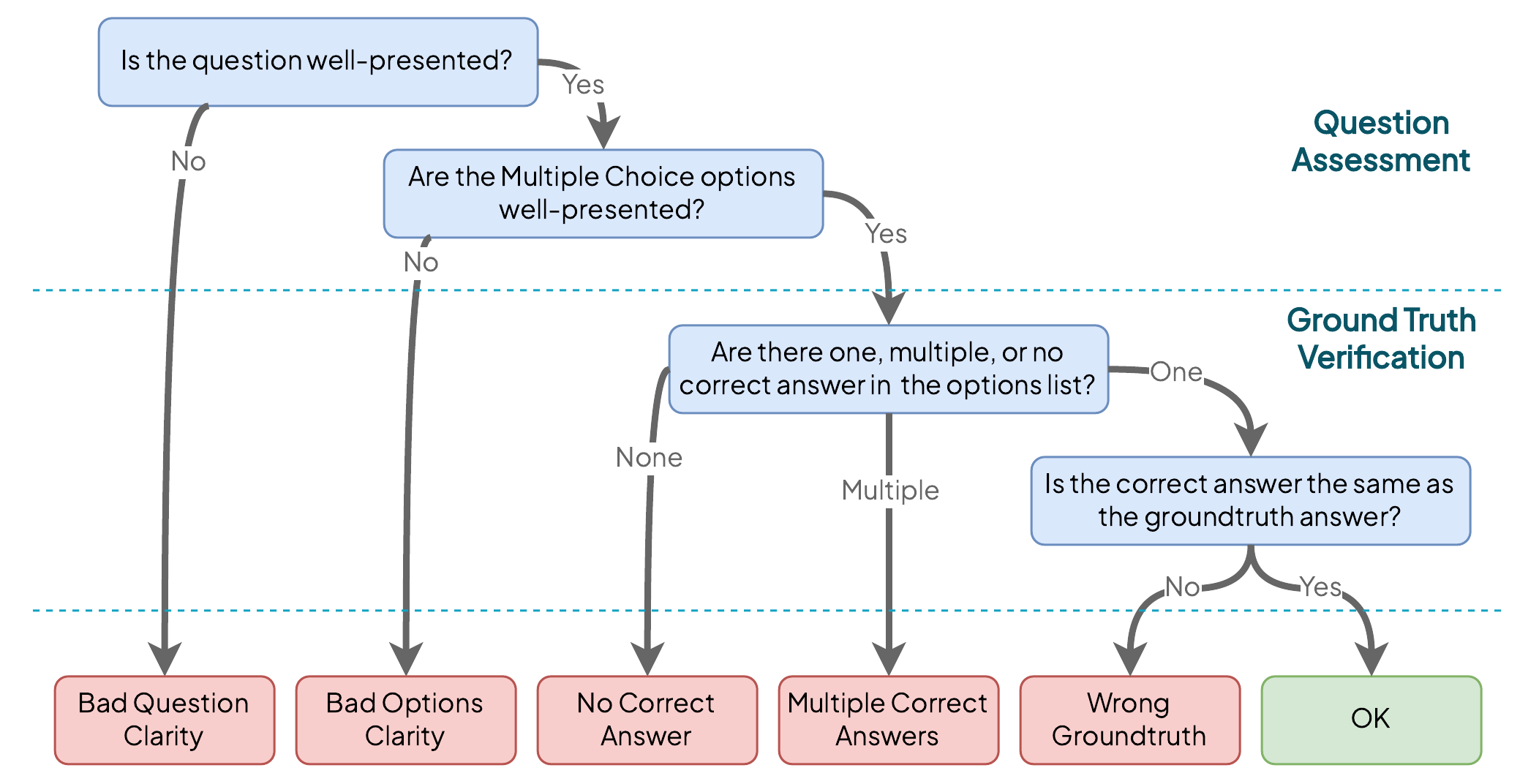}
    \caption{The error annotation protocol used for annotating instances in MMLU --- annotators first assess whether the instance is unambiguous (\emph{Question Assessment}, then whether there is a single valid answer, and then whether it matches with the ground truth answer from the dataset.}
    \label{fig:taxonomy}
\end{figure*}

\begin{figure}[t]
    \centering
        \centering
        \includegraphics[width=\linewidth]{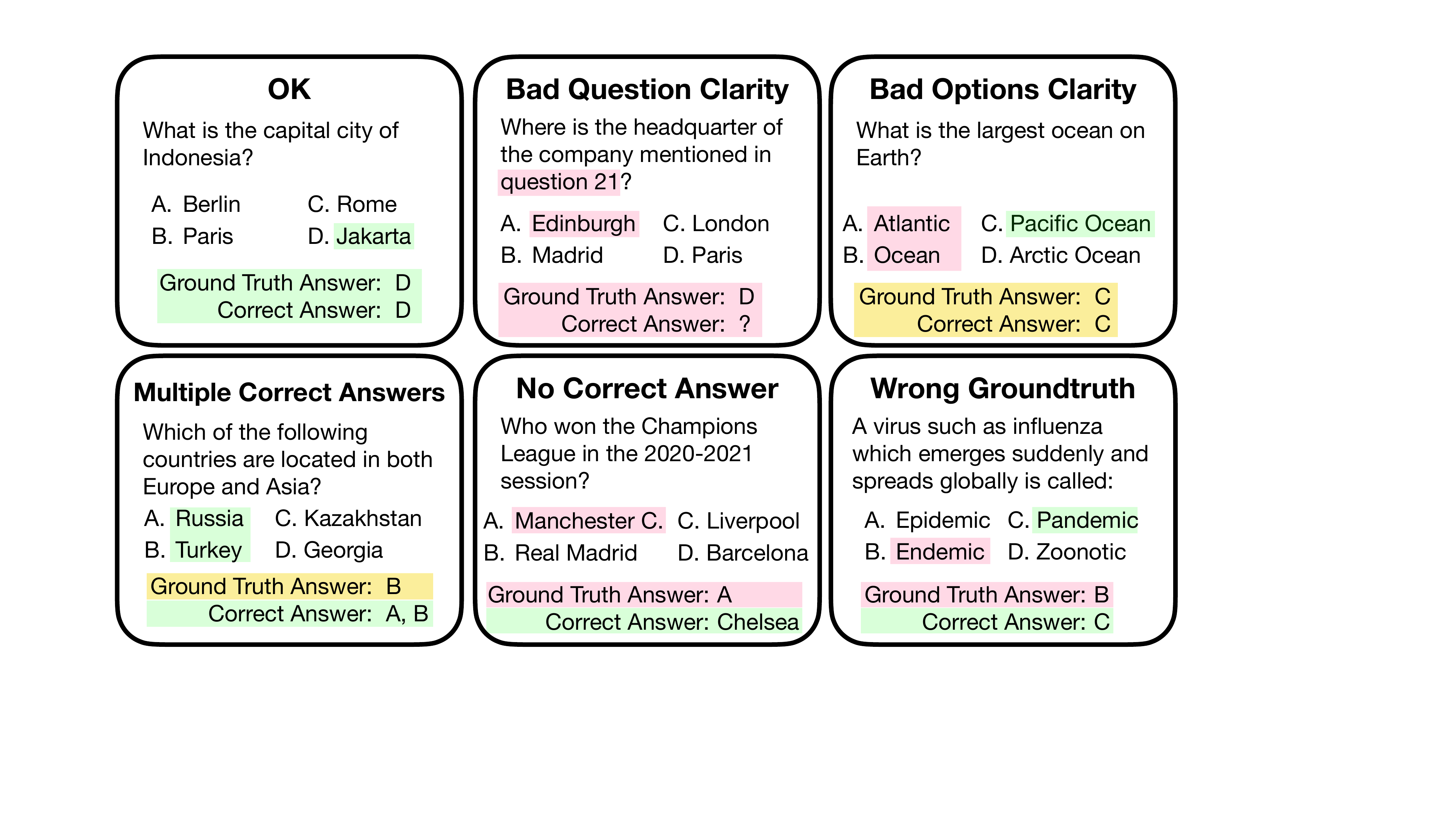}
        \caption{An example for every leaf node in the error annotation protocol shown in \cref{fig:taxonomy} from the MMLU dataset.}
        \label{fig:taxonomy-examples}
\end{figure}

We develop a hierarchical annotation protocol to classify the various errors identified in MMLU into specific error types. 
\Cref{fig:taxonomy} illustrates our annotation protocol for categorising MMLU errors, while \Cref{fig:taxonomy-examples} provides examples of each error category. 
We categorise errors into two primary groups: samples with errors in the clarity of the questions (Type 1, Question Assessment) and samples with errors in the ground truth answer (Type 2, Ground Truth Verification).

\textbf{(1a) Bad Question Clarity:} The question is poorly presented in terms of various aspects, such as clarity, grammar, and sufficiency of information. For instance, referring to a previous question.

\textbf{(1b) Bad Options Clarity:} The options are unclear, similar, or irrelevant to the question. Most errors in this category stem from incorrect parsing of the options from the original source. 
For example, a single option might be incorrectly split into two separate options. 

\textbf{(2a) No Correct Answer:} None of the options correctly answer the question.
This error may arise when the ground-truth options are omitted to reduce the number of options from five to four.
 
\textbf{(2b) Multiple Correct Answers:}  More than one option can be selected as the answer to the question. 
For example, the options contain a synonym of the ground truth label.
 
\textbf{(2c) Wrong Ground Truth:} The correct answer differs from the ground truth provided in MMLU.
This type of error occurs when the annotated label differs from the correct label, which may be caused by a mistake during manual annotation.
 
To create \dataset, we followed the proposed annotation protocol.
We aim to ensure comprehensive coverage of the different error types for further experiments.

\subsection{Heterogeneity of Errors in MMLU}

Our annotation process reveals substantial variation in error types across subsets. Some subsets are affected by ambiguous questions, while others by incorrect ground truth labels. These differences may impact how MMLU results are interpreted and addressed. Notable irregularities in certain subsets include (a full list of subsets and their corresponding errors can be found in \cref{sec: Appendix-B}):

\begin{description}[leftmargin=0pt,noitemsep,nolistsep]
\item[Virology] -- Incorrect ground truth labels are prevalent within the Virology subset. Many of the incorrect labels are for relatively simple questions, such as identifying the description of a pandemic; this suggests errors may stem from problems parsing the original datasets (in most cases, the \textit{Human Virology} textbook).
\item[College Chemistry] -- 
The questions were sourced from textbooks and standardised college-level exams. We identified erroneous questions resulting from simple parsing errors and unknown causes. For example, questions spanning multiple lines in the original source were often parsed incorrectly, leading to a part of the question being presented as the first option (Option A) and the exclusion of Option D. Furthermore, there were questions with ground truth labels that did not match the answers provided in the source without known cause.
\item[Professional Law] -- The benchmark does not clearly distinguish between different jurisdictions, despite focusing on U.S. law.
\item[Formal Logic] -- The dataset contains many questions with incorrect answers. These are mostly sourced from the `Oxford Learning Link' website. It is unknown what causes the inaccuracies. For example, one question states that $(F\wedge L)\wedge \neg C$ is correct, but $F\wedge L \wedge \neg C$ is not, even though these two formulas are clearly equivalent. 
\item[Global Facts] -- Almost all questions needed consulting external sources to be validated, where a large portion of them are reports from \url{ourworldindata.org} (18 cases) and \url{pewresearch.org} (15 cases); for several questions, multiple sources were providing conflicting answers --- for example, on the perceived corruption of political parties in 2013 Ethiopia, \url{ourworldindata.org} seems to confirm the answer in MMLU, while the Global Corruption Barometer from Transparency International was providing conflicting evidence.~\footnote{See data at \href{https://ourworldindata.org/grapher/share-of-people-who-think-political-parties-are-very-corrupt}{Our World in Data} and \href{https://images.transparencycdn.org/images/2013_GlobalCorruptionBarometer_EN_200525_112757.pdf}{Global Corruption Barometer}.}
\item[Machine Learning] -- Most questions were sourced from exam papers, assignments or online quizzes. About 30 of the questions require expert knowledge and reasoning. The main issue of this subset is the clarity of the questions and options. \eg some quiz questions are based on past knowledge, and the descriptions in the questions may be vague or inapplicable today. 
\item[Econometrics] -- The majority of the questions are correct, but some questions contain unverifiable references. \eg `Consider again the VAR model of equation 16,' but equation 16 cannot be found within the question.

\end{description}
These irregularities highlight error patterns in MMLU. We want to highlight one type of error that is especially challenging to catch, namely unspecified context that is needed to properly answer the question. For example, the Professional Law and Accounting datasets assume US jurisdiction and practices which are not specified in the question and may become outdated if standards change. Many subjects also display a US- and Western-centric bias: the Virology dataset refers to ``the Latino population'' in a US context, and the Human Aging dataset mentions an unspecified survey of older adults.

\begin{figure*}
  \begin{center}
    \includegraphics[width=\textwidth]{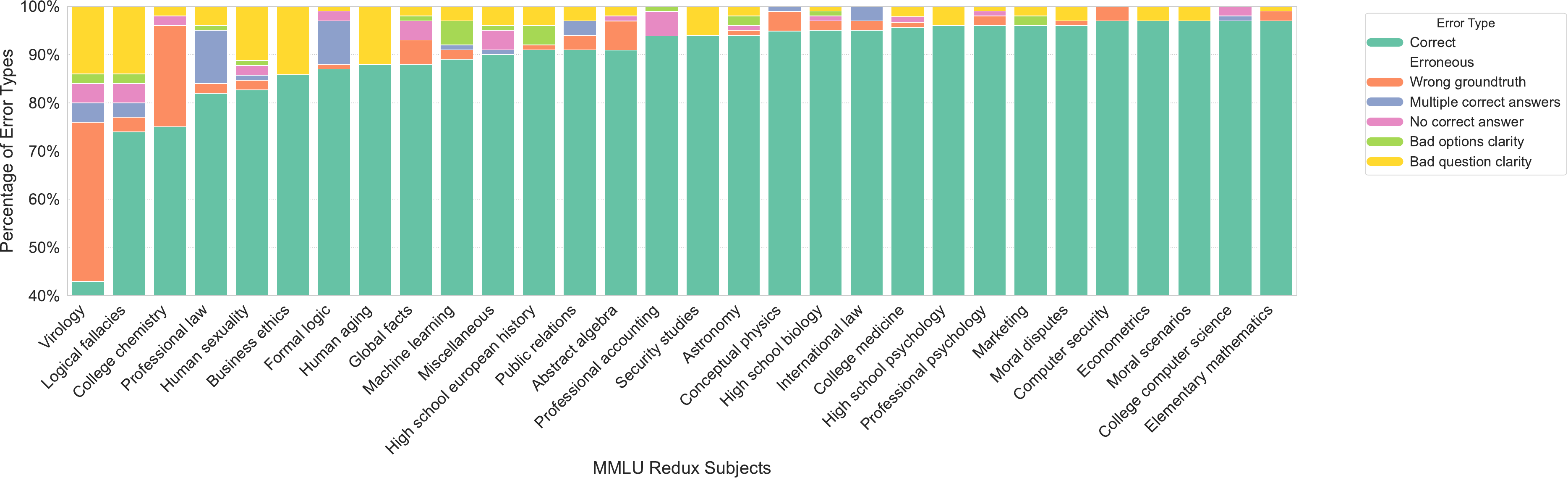}
  \end{center}
  \caption{Statistics of the error types detected in 100 randomly sampled and manually annotated data from the 30 most erroneous MMLU subjects. In the \emph{Virology} subject, we found that 57\% of the analysed instances contain some forms of errors, such as wrong ground truth (33\%), multiple correct answers (4\%), or unclear questions (14\%). For detailed numbers of all subsets, see Appendix \ref{app:mini-mmlu-stats-details}. 
  }
\label{fig:mmlu-redux-stats}
\end{figure*}

\section{\dataset: A Correct MMLU Subset}
\label{sec:dataset}

In this section, we propose \dataset, a manually annotated subset of MMLU, to quantify the errors present in the original dataset. \dataset serves two main purposes: \textit{1)} to measure the prevalence and types of errors in MMLU; and \textit{2)} to explore the feasibility of automatically fixing MMLU by leveraging the annotated error types. 
We find that \textit{1)} the proportion of errors in MMLU is non-negligible, highlighting the need for a correct subset; and \textit{2)} fixing MMLU automatically proves to be a challenging task, despite the availability of annotated error types.

We create \dataset by manually labelling a subset of MMLU questions with their corresponding error types. To this end, we follow the annotation protocol introduced in~\cref{sec:error_detection_taxonomy}. 
For more accurate annotations, we confirmed the error detection by finding the samples’ original source wherever it was available.
However, at present, the correct answers suggested by the annotators are not used to replace the existing MMLU labels.

In the following, we analyse the error statistics of \dataset and use \dataset to re-evaluate the performance of LLMs.
In \cref{sec:fix_mmlu_automatically}, we explore the possibility of using \dataset to improve the overall quality of MMLU by automatically fixing the identified errors.

\subsection{Analysis of \dataset}
\label{sec: MMLU-Redux}

We present the percentage of error types in \cref{fig:mmlu-redux-stats}, with detailed numbers available in \cref{app:mini-mmlu-stats-details}.
In our analysis, we find that more than 9\% of the examples are incorrect, suggesting a substantial presence of errors in the MMLU. 
Especially, we find that more than 57\% examples in Virology contain errors, where 30\% examples have a wrong ground truth, and 15\% are unclear questions.
Moreover, we also observe significant error percentages in other disciplines: more than 20\% examples in Logical Fallacies and College Chemistry are wrong, and more than 10\% examples in Professional Law, Business Ethics, Formal Logic, Human Aging, Global Facts, Machine Learning, Miscellaneous and Public Relations are wrong.
Such error proportions could lead to inaccurate comparisons and invalid rankings of LLM models.
Using stratified sampling, we estimate that 6.49\% of the questions in the complete MMLU dataset contain errors.  

\begin{figure*}
  \begin{center}
    \includegraphics[width=\textwidth]{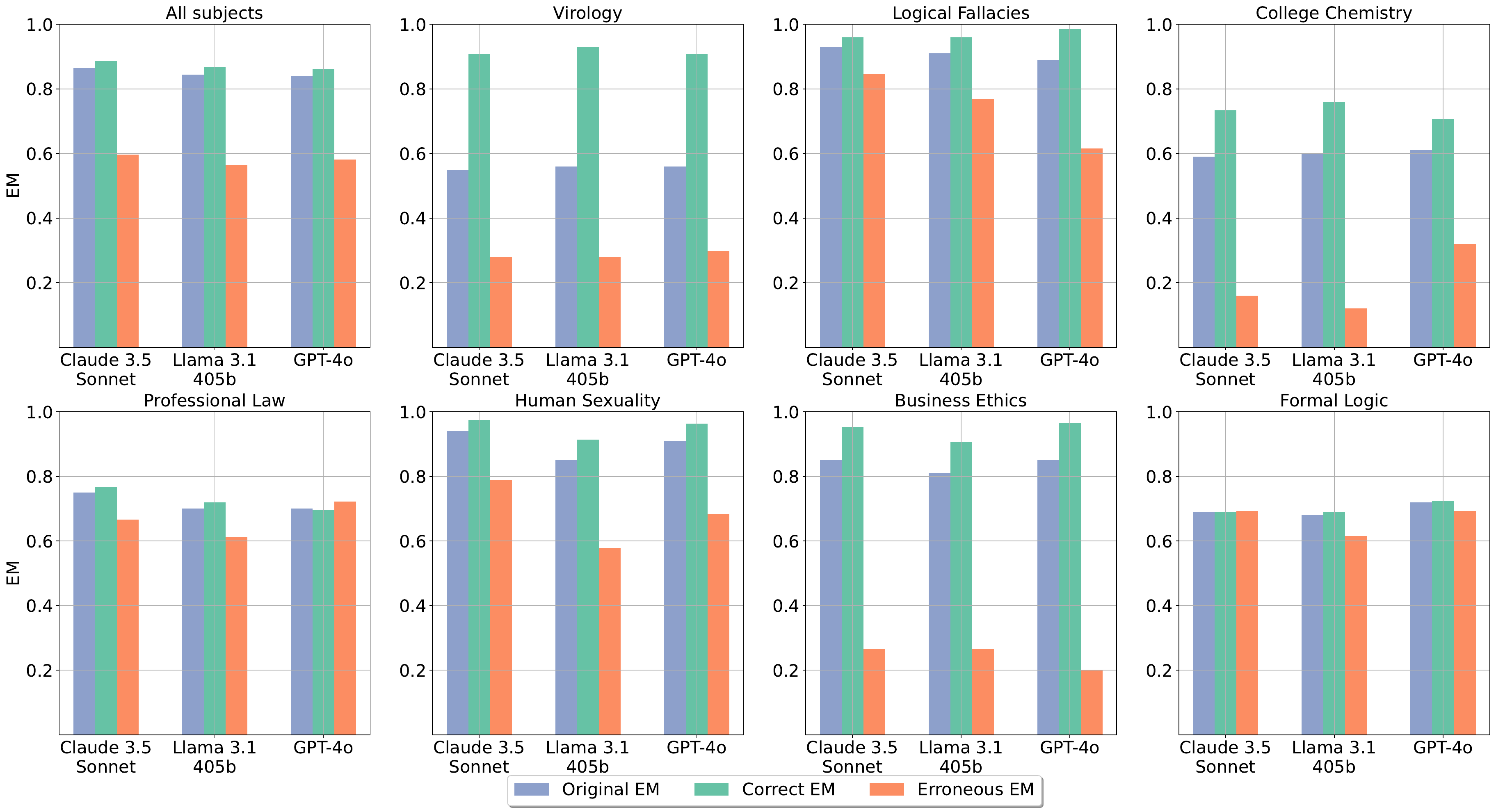}
  \end{center}
  \caption{Exact Match (EM) for each model on all subjects combined (top left), and the seven subjects with the most errors in our MMLU-Redux data. Original EM represents performance measured across all instances regardless of error presence, while Correct and Erroneous EM reflect results on correct instances and error instances, respectively. 
  }
\label{fig:mini-mmlu-em-error}
\end{figure*}

To better understand how these errors impact the performance of models on the MMLU, we compare the performance between erroneous instances and correct instances across the seven subjects identified as having the most errors (Virology, Logical Fallacies, College Chemistry, Professional Law, Business Ethics, Formal Logic, and Human Aging) in \cref{fig:mini-mmlu-em-error}. 

Despite the general trend indicating a performance decline among erroneous instances, we also observed cases where performance was similar or even higher in erroneous instances (Professional Law and Formal Logic). Given that LLMs should be unable to yield answers that are correct with respect to the original MMLU ground truth on these erroneous instances, this may be evidence of memorisation, suggesting that these MMLU instances were learned during the pretraining.

\begin{table}[t]
\caption{Inter-annotator agreement (Cohen's Kappa) analyses for the five erroneous subjects of MMLU-Redux. ``All Error Type'' is calculated based on whether the specific error types match, while ``Binary Error Type'' is calculated based on whether an error is present or not.}
  \label{table:agreement}
  \centering
  \resizebox{\linewidth}{!}{
\begin{tabular}{lcc}
\toprule
\textbf{Subject} &
  \textbf{\begin{tabular}[c]{@{}c@{}}Cohen's Kappa \\ (All Error Type)\end{tabular}} &
  \textbf{\begin{tabular}[c]{@{}c@{}}Cohen's Kappa \\ (Binary Error Type)\end{tabular}} \\
\midrule
Virology          & 0.67 & 0.67 \\
Logical Fallacies & 0.73 & 0.71 \\
College Chemistry & 0.92 & 0.95 \\
Formal Logic      & 0.96 & 1.0  \\
Human Sexuality   & 0.64 & 0.64 \\
\bottomrule
\end{tabular}
}
\end{table}

We also assessed the reliability of the manual annotations in MMLU-Redux by calculating Cohen's Kappa \citep{cohen1960coefficient} among three annotators for the five erroneous subjects (virology, logical fallacies, college chemistry, formal logic, and human sexuality). The results shown in \cref{table:agreement} demonstrate a generally high level of agreement across most subjects, with Cohen's Kappa values exceeding 0.6 for both ``All Error Type'' (whether the specific error types match) and ``Binary Error Type'' (whether an error is present or not). Additionally, the minimal difference in scores between ``All Error Type'' and ``Binary Error Type'' indicates that our error annotation protocol used for annotating (\cref{sec:error_detection_taxonomy}) is highly effective.

\subsection{Re-Evaluating the State-of-the-Art LLMs}

To assess how the corrected dataset impacts the performance of existing state-of-the-art LLMs, we re-evaluate them on the five subjects with the highest number of errors (Virology, Logical Fallacies, College Chemistry, Professional Law, and Human Sexuality) in \dataset.

In \cref{combined-rank-change-table}, we compared the performance of models when using all instances of \dataset to the performance when using only correct instances without errors to see if there are any changes in the rankings due to this.
The results clearly demonstrate that, at least for subjects with a high number of detected errors, these issues are significant enough to affect the results, leading to changes in model rankings.
For example, in the Virology subset, Llama 3.1 Instruct Turbo (405B) ranked 16th when considering all Virology instances, but ranked first when only correct instances were used.
On the other hand, in the Human Sexuality subset, GPT-4 (0613) achieved an exact match score of 0.91 across all instances (ranking 5th). However, when considering only the correct instances, the exact match score drops significantly to 0.43, placing it last among the top 10 models.
Since MMLU is an important benchmark for evaluating model performance, this indicates that errors in MMLU are a critical issue.

\begin{table*}[t]
    \caption{Comparison of model performance and ranking changes when using overall instances (as per HELM~\cite{liang2023holistic}) versus correct instances of \dataset on five most erroneous subjects. Numbers in parentheses denote ranks, \textcolor{teal}{teal} an improvement in rank, \textcolor{red}{red} a deterioration in rank, and \textcolor{black}{black} an unchanged rank.}
    \label{combined-rank-change-table}
    \centering
    \resizebox{\textwidth}{!}{
    \begin{tabular}{rccccccccccccc}

    \toprule
    \textbf{Model}                  & \textbf{Virology}                  & \textbf{Logical Fallacies}         & \textbf{College Chemistry}         & \textbf{Professional Law}         & \textbf{Human Sexuality}          \\
    \midrule
    Claude 3.5 Sonnet (20240620)    & \adjem{0.60}{1}{0.91}{5}           & \adjem{0.93}{1}{0.96}{5}           & \adjem{0.59}{9}{0.73}{4}           & \textbf{\adjem{0.75}{1}{0.77}{1}} & \textbf{\adjem{0.94}{1}{0.98}{1}} \\
    Claude 3 Opus (20240229)        & \adjem{0.58}{12}{0.88}{8}          & \adjem{0.90}{4}{0.96}{5}           & \adjem{0.60}{5}{0.72}{5}           & \adjem{0.72}{4}{0.72}{3}          & \adjem{0.91}{5}{0.96}{2}          \\
    Llama 3.1 Instruct Turbo (405B) & \textbf{\adjem{0.57}{16}{0.93}{1}} & \adjem{0.92}{2}{0.96}{5}           & \textbf{\adjem{0.60}{5}{0.76}{1}}  & \adjem{0.70}{6}{0.72}{3}          & \adjem{0.86}{20}{0.91}{9}         \\
    GPT-4o (2024-05-13)             & \adjem{0.60}{3}{0.91}{5}           & \adjem{0.88}{6}{0.99}{2}           & \adjem{0.61}{4}{0.71}{7}           & \adjem{0.72}{3}{0.70}{5}          & \adjem{0.91}{5}{0.96}{2}          \\
    Gemini 1.5 Pro (001)            & \adjem{0.55}{28}{0.91}{5}          & \adjem{0.90}{4}{0.99}{2}           & \adjem{0.62}{2}{0.72}{5}           & \adjem{0.67}{9}{0.67}{7}          & \adjem{0.37}{55}{0.94}{6}         \\
    GPT-4 (0613)                    & \adjem{0.60}{3}{0.86}{10}          & \adjem{0.87}{11}{0.99}{2}          & \adjem{0.55}{18}{0.75}{3}          & \adjem{0.73}{2}{0.68}{6}          & \adjem{0.91}{5}{0.43}{10}         \\
    Qwen2 Instruct (72B)            & \adjem{0.56}{24}{0.88}{8}          & \adjem{0.91}{3}{0.96}{5}           & \adjem{0.65}{1}{0.68}{8}           & \adjem{0.66}{10}{0.74}{2}         & \adjem{0.89}{11}{0.94}{6}         \\
    GPT-4 Turbo (2024-04-09)        & \textbf{\adjem{0.60}{1}{0.93}{1}}  & \textbf{\adjem{0.87}{11}{1.00}{1}} & \textbf{\adjem{0.53}{22}{0.76}{1}} & \adjem{0.67}{8}{0.63}{9}          & \adjem{0.90}{9}{0.93}{8}          \\
    Gemini 1.5 Pro (0409 preview)   & \textbf{\adjem{0.58}{10}{0.93}{1}} & \adjem{0.86}{18}{0.92}{10}         & \adjem{0.58}{13}{0.67}{9}          & \adjem{0.64}{13}{0.61}{10}        & \adjem{0.40}{56}{0.95}{5}         \\
    Llama 3.1 Instruct Turbo (70B)  & \textbf{\adjem{0.58}{12}{0.93}{1}} & \adjem{0.84}{27}{0.96}{5}          & \adjem{0.59}{9}{0.64}{10}          & \adjem{0.67}{7}{0.65}{8}          & \adjem{0.86}{20}{0.96}{2}         \\
    \bottomrule    
    \end{tabular}
    }
\end{table*}

\subsection{Correlation between Exact Match Score and Data Quality}

\begin{figure*}
  \begin{center}
    \includegraphics[width=\textwidth]{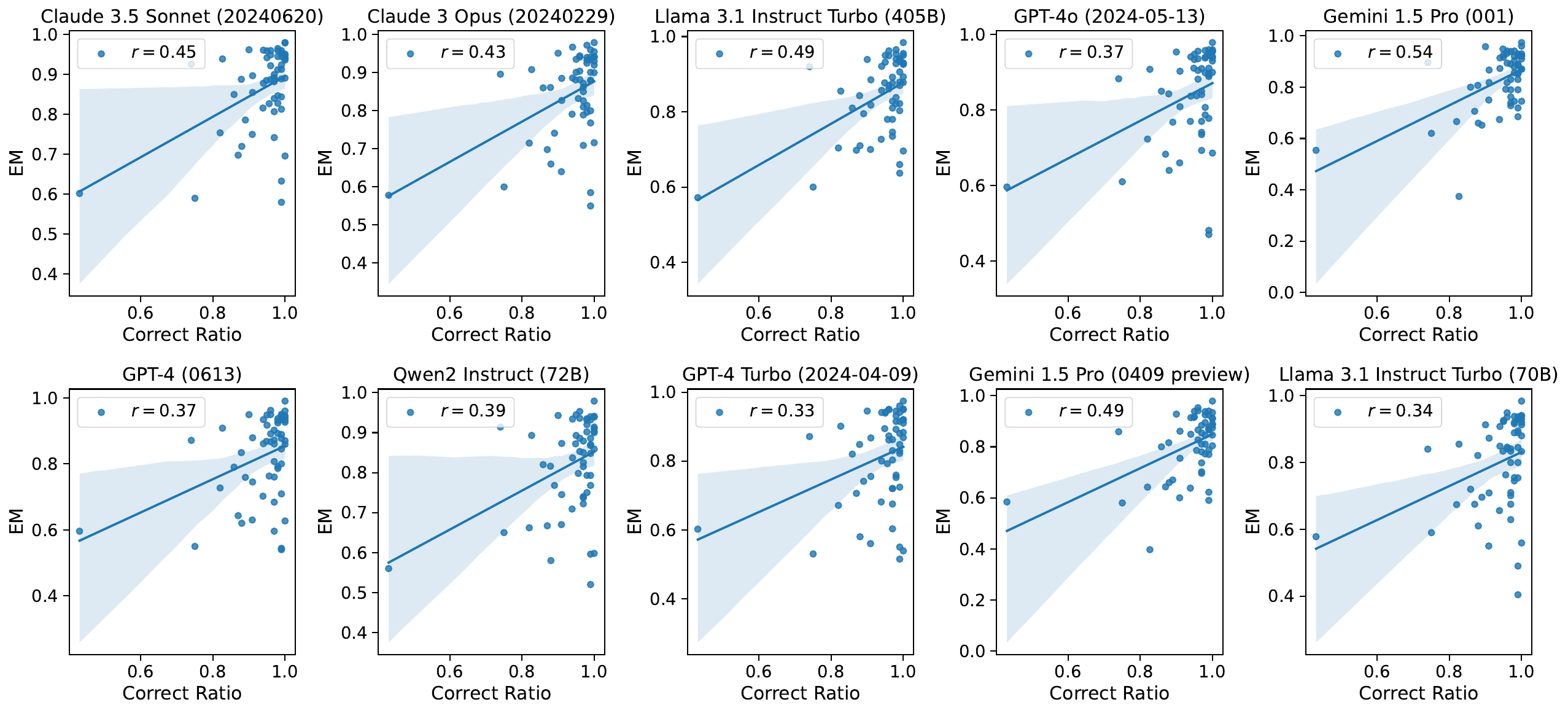}
  \end{center}
  \caption{Correlation between the performance of 10 best-performing LLMs on MMLU (as per HELM~\cite{liang2023holistic}) and the ratio of non-erroneous instances on MMLU-Redux. Each dot represents one MMLU subject.}
\label{fig:mmlu-redux-all-llms}
\end{figure*}

\Cref{fig:mmlu-redux-all-llms} shows the correlation between the performance of the ten best-performing LLMs on MMLU (as ranked by HELM~\cite{liang2023holistic}) and the ratio of non-erroneous instances on MMLU-Redux.
Each point in the scatter plot represents a subject in MMLU.
As expected, LLMs perform better in subjects with fewer errors, supporting the link between data reliability and model performance. Ideally, LLMs excel with non-erroneous instances and struggle with erroneous ones.

Outliers, however, reveal deviations from this trend, where models perform better or worse than expected for certain subjects.
These deviations suggest other factors influence performance, requiring further investigation.
For example, some LLMs perform well despite high error rates (in the Logical Fallacies subject), hinting at potential subject-specific advantages.
We further explore these anomalies and their implications for model robustness in \Cref{sec: MMLU-Redux}.

\section{Can We Fix the MMLU Dataset Automatically?}
\label{sec:fix_mmlu_automatically}
After presenting evidence of the numerous errors in the MMLU dataset, we explore the following approaches to detect these errors automatically: 

\begin{description}[leftmargin=0pt,noitemsep,nolistsep]
    \item[Zero-Shot prompting] We provide the model with a simple instruction to classify questions into ``ok'' or ``not ok'' without introducing any demonstrations. The prompt can be found in Appendix~\ref{standardprompting}.
    \item[Few-shot prompting]
    We provide the model with two examples for each error type to guide its classification decisions.
    \item[Chain of Thought (CoT) prompting]~\citep{wei2022chain} We encourage the model to generate reasoning steps before producing the final answer, in both zero-shot and few-shot settings. The prompt format can be found in Appendix~\ref{CoTprompting}.
    \item[Retrieval-augmented prompting (RAG)]
    We retrieve 5 most relevant paragraphs from Wikipedia and MS-MARCO and append them as context for zero-shot and CoT prompting. 
    \item[Instruction fine-tuning] We fine-tune Llama3 8B-Instruct model~\citep{llama3modelcard} using curated data and evaluate its performance on MMLU-Redux. Refer to Appendix \ref{sec: Appendix-E} for more details. 

\end{description}

In the following, we introduce these error detection strategies and their evaluation results in detail.

\subsection{Error Detection Experiments}

To evaluate the performance of LLMs in detecting errors in the \dataset dataset, we conduct experiments with 4 state-of-the-art models:  OpenAI's GPT-4 Turbo, GPT-4o, Anthropic's Claude-3 Opus, and Meta's LlamA-3-70B~\citep{llama3modelcard}. For each model, we test both standard prompting and Chain of Thought (CoT) prompting methods~\citep{wei2022chain}. Details about the prompts are provided in Appendix~\ref{standardprompting}.

\begin{table}[t]
    \caption{Comparison of different methods and models for error detection in \dataset scores.}
  \label{table:mmlu-scores-table}
  \centering
  \resizebox{\linewidth}{!}{
  \begin{tabular}{rcccc}
    \toprule
    \textbf{Model} & \textbf{Method} & \textbf{Recall} & \textbf{F1} & \textbf{F2} \\
    \midrule
    \multirow{4}{*}{GPT-4 Turbo} & Zero-Shot & 22.81 & 22.70 & 23.90 \\
     & Zero-Shot CoT & 27.97 & 22.97 & 23.47 \\
     & Few-shot & 26.55 & 27.93 & 27.97 \\
     & Few-shot CoT & 46.68 & 31.68 & 36.58 \\
    \midrule
    \multirow{4}{*}{GPT-4o} & Zero-Shot & 28.92 & 15.99 & 28.05 \\
     & Zero-Shot CoT & 19.85 & 21.41 & 19.98 \\
     & Few-shot & 37.40 & 24.93 & 31.06 \\
     & Few-shot CoT & 38.47 & 29.26 & 31.36 \\
    \midrule
    \multirow{4}{*}{Claude 3 Opus} & Zero-Shot & 27.11 & 27.00 & 27.65 \\
     & Zero-Shot CoT & 44.87 & \textbf{34.68} & 38.19 \\
     & Few-shot & 38.63 & 29.45 & 34.89 \\
     & Few-shot CoT & \textbf{48.85} & 24.03 & \textbf{40.29} \\
    \midrule
    \multirow{4}{*}{Llama3-70B} & Zero-Shot & 10.06 & 8.15 & 9.46 \\
     & Zero-Shot CoT & 10.74 & 8.15 & 10.17 \\
     & Few-shot & 17.82 & 18.58 & 17.60 \\
     & Few-shot CoT & 24.87 & 23.16 & 23.10 \\
    \bottomrule
  \end{tabular}
  }
\end{table}

We consider ``not ok'' as the positive class, and ``ok'' as the negative class for the calculation of Recall, F1, and F2 scores. 
Based on \cref{table:mmlu-scores-table}, the Few-shot CoT setting outperforms other settings across all models, suggesting that providing a small set of labelled examples with step-by-step reasoning instructions can improve error detection performance. However, even the best-performing model, Claude-3-opus, only achieves an F2 Score of 40.29, highlighting the difficulty of this task.

We then use RAG prompting to investigate the impact of external knowledge on error detection.
We use BM25 to retrieve five most relevant paragraphs from \textit{enwiki-paragraphs} and \textit{msmarco-v1-passage} corpus provided by Pyserini ~\citep{Lin_etal_SIGIR2021_Pyserini} using the question as the query. 
We then use these paragraphs as additional context in the prompt to classify the question and answer choices. 

\begin{table}[t]
    \caption{Comparison of different methods and models for error detection using RAG.}
  \label{table:mmlu-scores-table-rag}
  \centering
  \resizebox{\linewidth}{!}{
  \begin{tabular}{rccccc}
    \toprule
    \textbf{Model} & \textbf{Index} & \textbf{Method} & \textbf{Recall} & \textbf{F1} & \textbf{F2} \\
    \midrule
    \multirow{4}{*}{GPT-4 Turbo} & Wikipedia & Zero-Shot & 57.00 & 27.47 & 36.87 \\
     & Wikipedia & Zero-Shot CoT & 46.63 & \textbf{31.52} & 36.13 \\
     & MS MARCO & Zero-Shot & 57.11 & 26.34 & 35.02 \\
     & MS MARCO & Zero-Shot CoT & 37.20 & 26.07 & 29.64 \\
   
    \midrule
    \multirow{4}{*}{GPT-4o} & Wikipedia & Zero Shot & 35.01 & 27.25 & 29.79 \\
     & Wikipedia & Zero-Shot CoT & 33.67 & 26.90 & 28.75 \\
     & MS MARCO & Zero-Shot & 29.97 & 28.07 & 28.31 \\
     & MS MARCO & Zero-Shot CoT & 28.65 & 24.14 & 25.30 \\
    \midrule
    \multirow{4}{*}{Llama3-70B} & Wikipedia & Zero Shot & 22.78 & 20.41 & 21.00 \\
     & Wikipedia & Zero-Shot CoT & 10.12 & 12.61 & 10.90 \\
     & MS MARCO & Zero-Shot & 28.45 & 22.39 & 24.67 \\
     & MS MARCO & Zero-Shot CoT & 10.18 & 14.00 & 11.40 \\
    \midrule
    \multirow{2}{*}{Claude 3 Opus} & Wikipedia & Zero-Shot & 82.61 & 28.72 & \textbf{41.92} \\
     & MS MARCO & Zero-Shot &  \textbf{83.91} & 28.09 & 41.27 \\
   \bottomrule
  \end{tabular}
  }
\end{table}

Based on Table~\ref{table:mmlu-scores-table-rag}, the Claude 3 Opus model with the zero-shot method on the MS MARCO index achieves the highest Recall of 83.91. Claude 3 Opus achieves an F2 Score of 41.92 with the zero-shot method on the Wikipedia index. The GPT-4 models perform relatively worse than Claude, with the GPT-4o model showing lower scores than GPT-4 Turbo. Comparing the retrieval indexes, Wikipedia generally outperforms MS MARCO for the GPT-4o model, while the results are mixed for the GPT-4 Turbo and Llama3-70B models. The RAG approach outperforms the few-shot CoT setting mentioned in the previous analysis, indicating that incorporating retrieved information can enhance error detection performance.

Based on the results presented in Table~\ref{table:mmlu-scores-table} and~\ref{table:mmlu-scores-table-rag}, we can conclude that automatic error detection in the MMLU dataset remains a challenging task, despite the availability of annotated error types. While Claude 3 Opus demonstrates the highest Recall, F1, and F2 Scores compared to other models, Claude 3 Opus shows the highest performance with RAG, indicating its potential for identifying errors more effectively. However, the best-performing model and method combination using RAG still achieves relatively low scores, suggesting the overall reliability of the models in detecting errors across the diverse range of subjects in MMLU is still limited. Detailed performance across all subjects can be found in~\cref{sec: Appendix-C}.

\section{Related Work}

\paragraph{Benchmark Issues}
While benchmarks often enable methodological progress, they can also be counterproductive when labelling mistakes and annotation artefacts exist.
For example, \citet{DBLP:journals/corr/abs-2006-07159} show how annotation artefacts within the popular ImageNet benchmark \citep{russakovsky2015imagenet} led to likely overstated performance gains that did not necessarily transfer to other datasets and tasks.
In NLP, similar issues have been found in summarisation \citep{DBLP:conf/acl/TejaswinNL21,DBLP:conf/emnlp/BhandariGALN20} and natural language inference~\citep{DBLP:conf/naacl/GururanganSLSBS18,DBLP:conf/starsem/PoliakNHRD18,DBLP:conf/emnlp/StaceyMDRR20,DBLP:conf/acl/Wu0SD22} benchmarks.

Benchmark issues can arise from biases in the framing of the task~\citep{DBLP:conf/conll/SchwartzSKZCS17}; noisy annotations~\citep{DBLP:conf/acl/ChenBM16}; web-crawls \citep{raffel2023exploring,lee2021deduplicating,DBLP:journals/corr/abs-2304-08442}; automated labelling processes such as crowdsourcing \citep{DBLP:conf/socialcom/YuenKL11} --- where annotators may be optimising their own incentives; human errors~\citep{DBLP:conf/iccv/PetersonBGR19} --- \eg lack of expertise on a given topic; or programmatic weak supervision \citep{DBLP:journals/corr/abs-2202-05433,DBLP:conf/amia/GoswamiBD21,DBLP:journals/corr/abs-2307-10169}, and personal~\citep{DBLP:conf/emnlp/GevaGB19} or group-level~\citep{DBLP:conf/acl/LiuTMTYC0T22} annotator biases. 

\paragraph{MMLU Issues and MMLU-Pro.} The broad adoption of the MMLU benchmark for LLM evaluations~\citep{DBLP:journals/corr/abs-2312-11805, DBLP:journals/corr/abs-2303-08774, DBLP:journals/corr/abs-2307-09288, anthropic2024claude, llama3modelcard} means identifying issues or improvements is crucial for ensuring its continued applicability. 
Recent studies have identified issues with labelling errors and ambiguous questions in similar benchmarks, such as MedQA~\citep{DBLP:journals/corr/abs-2404-18416}.
Concurrent work developing the MMLU-Pro~\citep{wang2024mmlupro} benchmark also identifies a number of issues within a filtered and augmented subset of the original MMLU dataset and re-annotates parts of the dataset for inclusion in the new MMLU-Pro evaluation.
However, despite these efforts, errors from the original MMLU persist in the extended MMLU-Pro benchmark, highlighting the importance of our work in systematically identifying and addressing dataset errors. 
The recent Global MMLU dataset~\cite{singh2024globalmmluunderstandingaddressing} reevaluated MMLU to study which questions are culturally biased. 
We observed a similar Western bias in our qualitative analysis of the errors. 
Additionally, there is currently limited literature categorising and quantifying these issues across the original MMLU dataset to help inform our understanding of previous results. 
The AI Explained YouTube Channel recently highlighted several erroneous instances in MMLU across several subjects while evaluating their SmartGPT framework in a recent popular video~\citep{aiexplained}.
However, our study aims to provide a more systematic assessment of MMLU. 

\section{Conclusion} \label{sec:conclusions}

We analyse the MMLU benchmark, driven by the necessity for rigorous evaluation of its reliability. 
Our analysis using a hierarchical annotation protocol %
reveals that a significant portion of MMLU instances contain inaccuracies that could lead to misleading evaluation results.
For example, 57\% of the instances in the \emph{Virology} and 26\% in the \emph{Logical fallacies} subsets were found to be inaccurate.
To this end, we introduce \dataset, a thoroughly reviewed subset of the MMLU dataset \citep{DBLP:conf/iclr/HendrycksBBZMSS21} comprising 5,700 questions spanning all 57 MMLU subjects, and estimate that MMLU has an error rate of 6.49\%.
The re-evaluation of LLMs using \dataset shows a significant variation in performance metrics and shifts in model rankings for several subsets, emphasising the impact that dataset quality can have on the evaluation of LLMs.
Furthermore, we analyse whether it is possible to identify errors automatically; although Claude 3 Opus seems to produce the most accurate results on this task (41.9\% F2 score when using retrieval-augmented generation), it is still insufficient to produce a high-quality dataset.

\section*{Limitations}
\label{sec:limitation}
Although our analysis uses a significant subset of 5,700 questions, the accuracy of this analysis, both quantitatively and qualitatively, can be further improved with additional annotation of the remaining 8,342 questions in MMLU. 
Therefore, we open up \dataset for additional annotation of both additional MMLU subjects and add to the already reviewed subsets.
However, we acknowledge that the annotation protocol we introduced to classify errors might still be prone to the personal biases of the annotators.

\bibliography{references}

\onecolumn

\appendix

\clearpage

\section{Responsible Research Checklist}

\begin{enumerate}

\item For all authors...
\begin{enumerate}
  \item Do the main claims made in the abstract and introduction accurately reflect the paper's contributions and scope?
    \answerYes{} We claim to develop a dataset for studying the errors present in MMLU, which we present.
  \item Did you describe the limitations of your work?
    \answerYes{See Section \ref{sec:limitation}}
  \item Did you discuss any potential negative societal impacts of your work?
    \answerNo{See section \ref{sec:limitation}}
  \item Have you read the ethics review guidelines and ensured that your paper conforms to them?
    \answerYes{}
\end{enumerate}

\item If you are including theoretical results...
\begin{enumerate}
  \item Did you state the full set of assumptions of all theoretical results?
    \answerNA{}
	\item Did you include complete proofs of all theoretical results?
    \answerNA{}
\end{enumerate}

\item If you ran experiments (e.g. for benchmarks)...
\begin{enumerate}
  \item Did you include the code, data, and instructions needed to reproduce the main experimental results (either in the supplemental material or as a URL)?
    \answerYes{See \cref{sec:dataset}}
  \item Did you specify all the training details (e.g., data splits, hyperparameters, how they were chosen)? 
    \answerYes{} We provide details for all proposed methods. For prompting strategies, we show prompts in \cref{sec: Appendix-B} For fine-tuning, we provide details about the training dataset in \cref{sec: Appendix-E}.
	\item Did you report error bars (e.g., with respect to the random seed after running experiments multiple times)?
    \answerNo{} Rerunning experiments multiple times to obtain error bars would have exceeded our funding capabilities.
	\item Did you include the total amount of compute and the type of resources used (e.g., type of GPUs, internal cluster, or cloud provider)?
    \answerYes{} We provide details about our resources in \cref{sec: Appendix-E}
\end{enumerate}

\item If you are using existing assets (e.g., code, data, models) or curating/releasing new assets...
\begin{enumerate}
  \item If your work uses existing assets, did you cite the creators?
    \answerYes{} Our work is based on the MMLU dataset, which we cited in \cref{sec:introduction}.
  \item Did you mention the license of the assets?
    \answerYes{} The license of the dataset is available on the dataset URL (CC-BY 4.0).
  \item Did you include any new assets either in the supplemental material or as a URL?
    \answerYes{} \dataset is provided as a supplementary material.
  \item Did you discuss whether and how consent was obtained from people whose data you're using/curating?
    \answerYes{} Our dataset is based on MMLU, which has an MIT license. The annotation work for \dataset is done by the authors of this paper, who all gave consent to use the data.
  \item Did you discuss whether the data you are using/curating contains personally identifiable information or offensive content?
    \answerNo{} The MMLU data does not use personal data as it is based on publicly available test questions, and so neither does \dataset. 
\end{enumerate}

\item If you used crowdsourcing or conducted research with human subjects...
\begin{enumerate}
  \item Did you include the full text of instructions given to participants and screenshots, if applicable?
    \answerNA{}
  \item Did you describe any potential participant risks, with links to Institutional Review Board (IRB) approvals, if applicable?
    \answerNA{}
  \item Did you include the estimated hourly wage paid to participants and the total amount spent on participant compensation?
    \answerNA{}
\end{enumerate}

\end{enumerate}

\section{Data Collection and Organisation}
\label{sec:data_coll}

The \dataset dataset is available at \url{https://huggingface.co/datasets/edinburgh-dawg/mmlu-redux-2.0} (\texttt{doi:10.57967/hf/2507}) with a CC-BY-4.0 licence. This link also contains a public datasheet. 
The data used 
to create \dataset was obtained from \texttt{cais/mmlu}~\footnote{\url{https://huggingface.co/datasets/cais/mmlu}}, which is also utilised in the `lm-eval-harness' framework~\citep{eval-harness}. To ensure uniformity of our results, the language model (LM) predictions used in our performance analyses were obtained from the Holistic Evaluation of Language Models (HELM) leaderboard v1.3.0, released on May 15th, 2024.

We randomly subsampled 100 questions per MMLU subject to be presented to the annotators.
The annotators are instructed to follow the introduced taxonomy by first assessing the question presentation, and then by verifying the ground truth MMLU label.
The annotators were encouraged to perform an exact match search using a search engine to find occurrences of the question and multiple-choice options from credible sources.
If the annotators found an exact match of the question-options pair, the annotators were asked to evaluate the answer provided by the source.
Regardless of whether a label was found in the source, and whether the MMLU label is the same or not, the annotators were asked to decide whether they would follow the label using their expertise.
In the cases where an exact match was not found, the annotators were asked to search for supporting evidence from trusted sources, such as government websites, textbooks, and/or other reputable organisations (\eg World Health Organisation (WHO)).
In cases where the annotators are still unsure, they were asked to annotate the question with ``Expert'', denoting that the question requires more expertise.
This annotated subset of MMLU is denoted as \dataset.

\dataset comprises subsampled test splits of the aforementioned thirty MMLU subsets (denoted as ``\texttt{config}'' in HF vocabulary).

Each data point in \dataset contains seven columns:
\begin{itemize}[leftmargin=*]
    \item \textbf{question} (string): The original MMLU question.
    \item \textbf{choices} (list of four strings): The original list of four choices associated with the question from the MMLU dataset.
    \item \textbf{answer} (integer): The MMLU ground truth label in the form of an array index between 0 and 3.
    \item \textbf{error\_type} (string): The annotated error\_type. The values can be one of the six error types proposed in the taxonomy (``ok'', ``bad\_question\_clarity'', ``bad\_options\_clarity'', ``no\_correct\_answer'', ``multiple\_correct\_answers'', ``wrong\_groundtruth'') and ``expert''.
    \item \textbf{source} (string): The potential source of the question.
    \item \textbf{correct\_answer} (string): In the case of ``no\_correct\_answer'' and ``wrong\_groundtruth'', the annotators can suggest the alternative correct answer.
    \item \textbf{potential\_reason} (string): A free text column for the annotators to note what they believe to have caused the error.
\end{itemize}
The question, choices, and answer columns are taken from the MMLU Hugging Face dataset (\texttt{cais/mmlu}).

To ensure reproducibility and facilitate further research, we will make \dataset publicly available on the Hugging Face (HF) platform in the Croissant format~\cite{10.1145/3650203.3663326}.%
This format provides a standardised and easily accessible structure for the dataset, allowing researchers and practitioners to readily utilise it for their own investigations and analyses.
The code to generate the data and analyses will be made available in the camera ready. %

\section{Usage Guidelines}

To reproduce our results or perform analyses similar to those presented in this study, the user may download the data and utilise all the columns.
\dataset contains both correct and erroneous instances, so the user should look at the value in column ``error\_type'' to filter samples based on the specific error type.

In those cases where the error is ``no\_correct\_answer'', ``multiple\_correct\_answers'' or ``wrong\_groundtruth'', the users may examine the suggested answer reported in the ``correct\_answer'' column.
The user should consider that the questions and the options reported are the same as those in the MMLU dataset, and they have not been modified, even when affected by bad clarity.

\section{Maintenance Plan}

\subsection{General Maintenance}

The datasets will be updated as needed to maintain accuracy, with announcements made for each update. These updates will be posted on the dataset's Hugging Face page.
Older versions will be preserved and documented using the git commit hash of the dataset repository.

\subsection{Contributing Guidelines}
\label{sec:contributing}

We welcome contributions from the research community to enhance and expand the \dataset dataset. Primarily, we foresee two types of contribution:

\paragraph{Additional data annotation}

Currently, \dataset covers 5,700 MMLU questions, which is a subset of the full MMLU dataset.
We encourage the community to help annotate more questions to cover the entire dataset.
Generally, we recommend you to follow these steps:

\begin{enumerate}[leftmargin=*]
    \item Familiarise yourself with the taxonomy. The taxonomy is designed to be simple and broad to cover all possible erroneous cases found in MMLU.
    \item Open a new discussion on the Hugging Face page of \dataset (\url{https://huggingface.co/datasets/edinburgh-dawg/MMLU-Redux/discussions/new})
    \item Insert the title with the prefix ``[ADD]'' to signify additional data annotation, followed by the name of the MMLU subject you are contributing to. For example, ``[ADD] Virology''.
    \item In the description body of the issue, specify the question, the choices, the MMLU answer, the error type, the correct answer (if applicable), the question source, what you believe to be the reason for the error, and optionally an additional comment. Follow this markdown template: 
\begin{lstlisting}
**Question**: {question}
**Choices**: {choices}  # Copy paste the python array
**MMLU answer**: {mmlu_answer}

**Error type**: {error_type}
# If this field is not applicable (e.g., for bad question clarity 
# or bad options clarity), write "N/A"
**Correct answer**: {correct_answer}  
# Ideally, a URL to where you can find the source online.
**Source**: {source}  
 # This is strictly to describe what might have caused the error 
 # (e.g., incorrect HTML parsing)
**Potential error reason**: {free text} 
# This is where the annotators can leave additional comments
**Additional comment**: {free text}  
\end{lstlisting}
    \item If you intend to propose annotation of more than one question, we ask you to use a delimiter such as three-dashes (``-{}-{}-'').
    \item Submit the discussion.
\end{enumerate}

We aim to respond to discussions within 48 hours.
The proposed changes will either be integrated into the associated MMLU subject or receive feedback for further adjustments and clarification.
Discussions with fully integrated changes will be closed.

\paragraph{Fix to the existing annotation} 

The current and future annotated questions in \dataset may contain inaccuracies, particularly for questions whose answers may change over time, and other factors.
We believe that the community can help keep the data up to date.
We recommend everyone to propose fixes by following these steps:

\begin{enumerate}[leftmargin=*]
    \item Familiarise yourself with the taxonomy. The taxonomy is simple and broad to cover all possible erroneous cases found in MMLU.
    \item Open a new discussion on the Hugging Face page of \dataset (\url{https://huggingface.co/datasets/edinburgh-dawg/MMLU-Redux/discussions/new})
    \item Insert the title with the prefix ``[FIX]'' to signify additional data annotation, followed by the name of the MMLU subject you are contributing to. For example, ``[FIX] Virology''.
    \item In the description body, specify the question, the choices, the MMLU answer, the current error type, the new error type, the current possible correct answer, the new correct answer, the question source, the potential reason that causes the error, and the optional additional comments. Follow this markdown template:
\begin{lstlisting}
**Question**: {question}
**Choices**: {choices}  # Copy paste the python array
**MMLU answer**: {mmlu_answer}

**Current Error type**: {current_error_type}
**Current Correct answer**: {current_correct_answer}
**New Error type**: {new_error_type}
# If it is not applicable 
# (e.g., for bad question clarity or bad options clarity), write "N/A"
**New Correct answer**: {new_correct_answer}  
# Ideally, a URL to where you can find the source online.
**Source**: {source}  
# This is strictly to describe what might have caused the error 
# (e.g., incorrect HTML parsing)
**Potential error reason**: {free text}  
# This is where the annotators can leave additional comments
# about the data
**Additional comment**: {free text}  
\end{lstlisting}
    \item If you intend to propose more than one fix, please use a delimiter such as three-dashes (``-{}-{}-'').
    \item Submit the discussion.
\end{enumerate}

For contributions that may not be suitable for submission via Hugging Face discussions or pull requests (e.g., bulk annotations stored in a CSV/PDF/DOCX file), we encourage you to reach out to us directly at 
\url{aryo.gema@ed.ac.uk} and \url{p.minervini@ed.ac.uk}, 
together with your HF username (if you have one and you are willing to be acknowledged). 
We will create a pull request which contains the proposed changes and tag your HF username. You can verify the changes and notify us via the discussion page if the pull request is ready to be merged.

\subsection{Contingency Plan}

We are committed to ensuring the long-term availability and accessibility of \dataset.
If the primary hosting platform (\ie Hugging Face) becomes unavailable or ceases operations, we will immediately activate our contingency measures. This includes notifying all stakeholders and users of the dataset about the temporary unavailability and redirecting them to the mirror copies available on alternative platforms (\ie Zenodo, 
\url{https://zenodo.org/records/14051942}).
If the primary hosting ceases operations, we will ensure the dataset is reuploaded to a new reliable hosting platform, accompanied by updated documentation and links to the new location. We will also utilise social media and other networks to inform the community about the new location of the dataset.

\section{Authors Statement}

The authors bear all responsibility in case of violation of rights.

\section{\dataset Error Type Statistics}

\label{app:mini-mmlu-stats-details}
In this section, we give the exact statistics of the error types found in \dataset. \cref{table:question-analysis} contains an overview. We include the total number of questions in each subject, but note that we annotated a subset of 100 questions for each. 
\begin{table}[htp]
\centering
\caption{Error types by subset. BQC is (1a) Bad Question Clarity, BOC is (1b) Bad Options Clarity, NCA is (2a) No Correct Answer, MCA is (2b) Multiple Correct Answers, and WGT is (2c) Wrong Ground Truth. The percentages in the Total row are estimated via stratified sampling over the subsets. }
\label{table:question-analysis}
\footnotesize
\begin{tabular}{lcccccccc}
\toprule
{\bf Subset} & {\bf Original Size} & {\bf Annotated} & {\bf OK} & {\bf BQC} & {\bf BOC} & {\bf NCA} & {\bf MCA} & {\bf WGT} \\
\midrule
Virology & 166 & 100 & 43 & 14 & 2 & 4 & 4 & 33 \\
Logical fallacies & 163 & 100 & 74 & 14 & 2 & 4 & 3 & 3 \\
College chemistry & 100 & 100 & 75 & 2 & 0 & 2 & 0 & 21 \\
Professional law & 1,534 & 100 & 82 & 4 & 1 & 0 & 11 & 2 \\
Human sexuality & 131 & 100 & 83 & 11 & 1 & 2 & 1 & 2 \\
Business ethics & 100 & 100 & 86 & 14 & 0 & 0 & 0 & 0 \\
Formal logic & 126 & 100 & 87 & 1 & 0 & 2 & 9 & 1 \\
Human aging & 223 & 100 & 88 & 12 & 0 & 0 & 0 & 0 \\
Global facts & 100 & 100 & 88 & 2 & 1 & 4 & 0 & 5 \\
Machine learning & 112 & 100 & 89 & 3 & 5 & 0 & 1 & 2 \\
Miscellaneous & 783 & 100 & 90 & 4 & 1 & 4 & 1 & 0 \\
Abstract algebra & 100 & 100 & 91 & 2 & 0 & 1 & 0 & 6 \\
Public relations & 110 & 100 & 91 & 3 & 0 & 0 & 3 & 3 \\
High school European history & 165 & 100 & 91 & 4 & 4 & 0 & 0 & 1 \\
Professional accounting & 282 & 100 & 94 & 0 & 1 & 5 & 0 & 0 \\
Astronomy & 152 & 100 & 94 & 2 & 2 & 1 & 0 & 1 \\
Security studies & 245 & 100 & 94 & 6 & 0 & 0 & 0 & 0 \\
Conceptual physics & 235 & 100 & 95 & 0 & 0 & 0 & 1 & 4 \\
International law & 121 & 100 & 95 & 0 & 0 & 0 & 3 & 2 \\
High school biology & 310 & 100 & 95 & 1 & 1 & 1 & 0 & 2 \\
College medicine & 173 & 100 & 96 & 2 & 0 & 1 & 0 & 1 \\
Moral disputes & 346 & 100 & 96 & 3 & 0 & 0 & 0 & 1 \\
Marketing & 234 & 100 & 96 & 2 & 2 & 0 & 0 & 0 \\
Professional psychology & 612 & 100 & 96 & 1 & 0 & 1 & 0 & 2 \\
High school psychology & 545 & 100 & 96 & 4 & 0 & 0 & 0 & 0 \\
High school physics & 151 & 100 & 97 & 0 & 0 & 1 & 0 & 2 \\
Elementary mathematics & 378 & 100 & 97 & 1 & 0 & 0 & 0 & 2 \\
High school macroeconomics & 390 & 100 & 97 & 2 & 0 & 0 & 1 & 0 \\
College computer science & 100 & 100 & 97 & 0 & 0 & 2 & 1 & 0 \\
Computer security & 100 & 100 & 97 & 0 & 0 & 0 & 0 & 3 \\
Moral scenarios & 895 & 100 & 97 & 3 & 0 & 0 & 0 & 0 \\
Econometrics & 114 & 100 & 97 & 3 & 0 & 0 & 0 & 0 \\
Nutrition & 306 & 100 & 98 & 1 & 1 & 0 & 0 & 0 \\
High school microeconomics & 238 & 100 & 98 & 1 & 0 & 0 & 0 & 1 \\
Sociology & 201 & 100 & 98 & 2 & 0 & 0 & 0 & 0 \\
High school statistics & 216 & 100 & 98 & 0 & 0 & 0 & 0 & 2 \\
College biology & 144 & 100 & 98 & 1 & 0 & 0 & 0 & 1 \\
Electrical engineering & 145 & 100 & 98 & 1 & 0 & 0 & 0 & 1 \\
High school world history & 237 & 100 & 99 & 0 & 0 & 0 & 0 & 1 \\
Anatomy & 135 & 100 & 99 & 0 & 0 & 0 & 0 & 1 \\
High school mathematics & 270 & 100 & 99 & 0 & 0 & 0 & 0 & 1 \\
High school chemistry & 203 & 100 & 99 & 1 & 0 & 0 & 0 & 0 \\
Jurisprudence & 108 & 100 & 99 & 0 & 1 & 0 & 0 & 0 \\
World religions & 171 & 100 & 99 & 1 & 0 & 0 & 0 & 0 \\
US foreign policy & 100 & 100 & 99 & 1 & 0 & 0 & 0 & 0 \\
Clinical knowledge & 265 & 100 & 99 & 1 & 0 & 0 & 0 & 0 \\
Professional medicine & 272 & 100 & 99 & 1 & 0 & 0 & 0 & 0 \\
College mathematics & 100 & 100 & 99 & 1 & 0 & 0 & 0 & 0 \\
Management & 103 & 100 & 100 & 0 & 0 & 0 & 0 & 0 \\
Philosophy & 311 & 100 & 100 & 0 & 0 & 0 & 0 & 0 \\
Medical genetics & 100 & 100 & 100 & 0 & 0 & 0 & 0 & 0 \\
High school US history & 204 & 100 & 100 & 0 & 0 & 0 & 0 & 0 \\
High school government and politics & 193 & 100 & 100 & 0 & 0 & 0 & 0 & 0 \\
High school geography & 198 & 100 & 100 & 0 & 0 & 0 & 0 & 0 \\
High school computer science & 100 & 100 & 100 & 0 & 0 & 0 & 0 & 0 \\
College physics & 102 & 100 & 100 & 0 & 0 & 0 & 0 & 0 \\
Prehistory & 324 & 100 & 100 & 0 & 0 & 0 & 0 & 0 \\
\midrule
Total & 14,042 & 5,700 & 93.51\% & 2.47\% & 0.44\% & 0.62\% & 1.54\% & 1.42\% \\
\bottomrule
\end{tabular}
\end{table}

\clearpage
\section{Prompting Methods}
Below, we provide the prompts used for both standard prompting and Chain of Thought (CoT) prompting methods.

Throughout the evaluation, we used the test split of the \dataset loaded with the specified configuration. The hyperparameters include a temperature of 0.0, top\_p of 1, frequency\_penalty and presence\_penalty of 0, and max\_tokens of 600 for both the standard prompting and Chain of Thought (CoT) prompting methods to ensure consistency and deterministic results. The default random seed was used.

\begin{tcolorbox}
[colback=gray!25!white,colframe=gray!100!gray,colbacktitle=gray!90!white, title=Standard Prompting,fonttitle=\bfseries]
\label{standardprompting}
\small
\# Task:\\
Given a question, its choices, and the ground truth answer, classify the question as either `ok' or `not ok'.\\

- `ok' means that the question and the choices are understandable, and the ground truth answer is correct.\\
- `not ok' means that the ground truth answer is incorrect, or the question and the choices are not well presented.\\
Classify with `ok' or `not ok' WITHOUT PROVIDING ANY REASONING
\end{tcolorbox}

\begin{tcolorbox}[colback=gray!25!white,colframe=gray!100!gray,colbacktitle=gray!90!white, title= Chain of Thought (CoT) Prompting,fonttitle=\bfseries]
\label{CoTprompting}
\small
\# Task:\\
Given a triple consisting of a multiple choice question, its choices, and the corresponding ground truth answer, your task is to classify the triple into `ok' or `not ok'.\\

\# Instructions:\\
1. Question Presentation: Is the question well-presented? Assess clarity, grammar, and sufficiency of information.\\
1.1 If Yes, assess the presentation of the MC options.\\
1.2 If No, classify the issue as `not ok'.\\
2. MC Options Presentation: Are the MC options well-presented? Check if the options are clear, distinct, and relevant to the question.\\
2.1 If Yes, determine if there is one potentially correct answer.\\
2.2 If No, classify the issue as `not ok'.\\
3. Answer Evaluation: Is there one, more than one, or no potentially correct answer in the options list?\\
3.1 If one, continue to Ground Truth Answer Evaluation.\\
3.2 If more than one, classify the issue as `not ok'.\\
3.3 If no correct answer, classify the issue as `not ok'.\\
4. Ground Truth Answer Evaluation: Is the ground truth answer correct?\\
4.1. If Yes, classify as `ok'.\\
4.2. If No, classify as `not ok'.\\
Provide your assessment in JSON format with keys `Question Presentation', `MC Options Presentation', `Answer Evaluation', `Ground Truth Answer Evaluation', `Classification'. \\
The `classification' is either `ok', or `not ok'.\\
FOLLOW THE EXACT EXAMPLE ANSWER FORMAT ALL IN ONE LINE WITHOUT PROVIDING EXPLANATION\\
\# Example Answer:\\
\{``Question Presentation'': ``Correct'', ``MC Options Presentation'': ``Correct'', ``Answer Evaluation'': ``One'', ``Ground Truth Answer Evaluation'': ``Correct'', ``Classification'': ``Correct''\}
\end{tcolorbox}

\clearpage

\section{Details on automatic error detection}
\label{sec: Appendix-C}

\subsection{Detailed Results on Error Detection Experiments for \dataset}

\begin{table}[htp]
  \caption{Comparison of Zero-Shot, Chain-of-Thought (CoT), and Few-Shot F1 score on 10 \dataset subjects.}
  \label{gpt-f1-score-table}
  \centering
  \scalebox{0.9}{
  \begin{tabular}{lccccc}
    \toprule
    \textbf{Dataset} & \textbf{Models}&{\bf Zero-Shot} & {\bf Zero-Shot CoT} & {\bf Few-Shot} & {\bf Few-Shot CoT} \\
    \midrule
    \multirow{4}{*}{College Chemistry}  & GPT-4-Turbo & 52.94 & 17.14 & 55.74 & 48.98 \\
     & GPT-4o & 42.86 & 30.77 & 19.35 & 30.77 \\
    & Claude-3-Opus & 40.00 & 36.36 & 24.24 & 32.26 \\
    & Llama-3-70B & 0.00 & 0.00 & 13.00 & 19.00 \\
    \midrule
    \multirow{4}{*}{College Mathematics}  & GPT-4-Turbo &0.00 & 20.00 & 0.00 & 0.00 \\
    &  GPT-4o &0.00 & 21.43 & 0.00 & 21.43 \\
    &Claude-3-Opus &  9.52 & 0.00 & 0.00 & 32.26 \\
    & Llama-3-70B & 0.00 & 0.00 & 0.00 & 0.00 \\
    \midrule
    \multirow{4}{*}{Econometrics} & GPT-4-Turbo &0.00 & 8.70 & 28.57 & 20.69 \\
    &  GPT-4o & 0.00 & 29.41 & 40.00 & 29.41 \\
    & Claude-3-Opus & 15.38 & 0.00 & 46.15 & 0.00 \\
    & Llama-3-70B & 0.00 & 0.00 & 57.00 & 55.00 \\
    \midrule
    \multirow{4}{*}{Formal Logic} & GPT-4-Turbo &23.33 & 22.22 & 26.09 & 0.00 \\
    &  GPT-4o & 33.33 & 0.00 & 0.00 & 0.00 \\
    & Claude-3-Opus & 26.67 & 55.81 & 0.00 & 0.00 \\
    & Llama-3-70B & 0.00 & 0.00 & 0.00 & 0.00 \\
    \midrule
    \multirow{4}{*}{Global Facts} & GPT-4-Turbo &19.35 & 30.00 & 20.00 & 44.44 \\
    &  GPT-4o & 16.67 & 28.57 & 21.05 & 28.57 \\
    & Claude-3-Opus & 28.57 & 32.26 & 38.09 & 37.50 \\
    & Llama-3-70B & 0.00 & 0.00 & 22.00 & 32.00 \\
    \midrule
    \multirow{4}{*}{High School Physics} & GPT-4-Turbo &9.52 & 17.39 & 23.53 & 16.67 \\
    &  GPT-4o & 19.05 & 26.09 & 30.77 & 26.09 \\
    & Claude-3-Opus & 8.33 & 21.05 & 28.57 & 33.33 \\
    & Llama-3-70B & 0.00 & 0.00 & 18.00 & 22.00 \\
    \midrule
    \multirow{4}{*}{Machine Learning} & GPT-4-Turbo &20.83 & 9.52 & 30.77 & 35.71 \\
    &  GPT-4o & 40.00 & 20.00 & 28.57 & 20.00 \\
    & Claude-3-Opus &  33.33 & 23.08 & 31.58 & 46.15 \\
    & Llama-3-70B & 0.00 & 0.00 & 0.00 & 14.00 \\
    \midrule
    \multirow{4}{*}{Professional Law} & GPT-4-Turbo &25.00 & 17.14 & 21.43 & 22.86 \\
    &  GPT-4o & 8.00 & 23.53 & 8.00 & 23.53 \\
    & Claude-3-Opus & 29.79 & 32.43 & 16.00& 23.53 \\
    & Llama-3-70B & 11.00 & 9.00 & 8.00 & 9.00 \\
    \midrule
    \multirow{4}{*}{Public Relations} & GPT-4-Turbo &0.00 & 20.00 & 31.58 & 52.17 \\
    &  GPT-4o & 0.00 & 31.58 & 25.00 & 31.58 \\
    & Claude-3-Opus & 0.00 & 80.77 & 38.09 & 35.29 \\
    & Llama-3-70B & 17.00 & 17.00 & 15.00 & 14.00 \\
    \midrule
    \multirow{4}{*}{Virology} & GPT-4-Turbo &76.00 & 73.12 & 81.19 & 75.27 \\
    &  GPT-4o & 0.00 & 81.19 & 76.59 & 81.19 \\
    & Claude-3-Opus & 78.43 & 25.00 & 71.74 & 0.00 \\
    & Llama-3-70B & 54.00 & 56.00 & 52.00 & 67.00 \\
    \bottomrule
  \end{tabular}
  }
\end{table}

\begin{table}[htp]
  \caption{Comparison of Zero-Shot, Chain-of-Thought (CoT), and Few-Shot F2 score on 10 \dataset subjects.}
  \label{gpt-f2-score-table}
  \centering
  \scalebox{0.9}{
  \begin{tabular}{lccccc}
    \toprule
    \textbf{Dataset} & \textbf{Models}&{\bf Zero-Shot} & {\bf Zero-Shot CoT} & {\bf Few-Shot} & {\bf Few-Shot CoT} \\
    \midrule
    \multirow{4}{*}{College Chemistry}  & GPT-4-Turbo & 45.69 & 26.55 & 50.29 & 48.39 \\
     & GPT-4o & 48.39 & 18.69 & 30.61 & 22.94 \\
    & Claude-3-Opus & 40.00 & 50.85 & 35.09 & 27.27 \\
    & Llama-3-70B & 0.00 & 0.00 & 0.00 &  0.00 \\
    \midrule
    \multirow{4}{*}{College Mathematics}  & GPT-4-Turbo &0.00 & 0.00 & 0.00 & 0.00 \\
    &  GPT-4o &0.00 & 0.00 & 0.00 & 0.00 \\
    &Claude-3-Opus &  6.17 & 0.00 & 0.00 & 0.00 \\
    & Llama-3-70B & 0.00 & 0.00 & 0.00 & 10.64 \\
    \midrule
    \multirow{4}{*}{Econometrics} & GPT-4-Turbo &6.13 & 15.63 & 10.49 & 39.47\\
    &  GPT-4o & 13.33 & 0.00 & 32.26 & 50.00 \\
    & Claude-3-Opus & 10.53 & 40.00 & 34.88 & 60.00 \\
    & Llama-3-70B & 0.00 & 0.00 & 9.52 & 14.15 \\
    \midrule
    \multirow{4}{*}{Formal Logic} & GPT-4-Turbo &17.41 & 22.73 & 26.32 & 20.83 \\
    &  GPT-4o & 35.09 & 8.47 & 30.30 & 24.19 \\
    & Claude-3-Opus & 24.69 & 32.79 & 40.54 & 31.75 \\
    & Llama-3-70B & 0.00 & 0.00 & 0.00 & 0.00 \\
    \midrule
    \multirow{4}{*}{Global Facts} & GPT-4-Turbo &17.05 & 26.79 & 25.00 & 37.04 \\
    &  GPT-4o & 16.67 & 9.09 & 25.00 & 17.86 \\
    & Claude-3-Opus & 31.25 & 37.31 & 41.67 & 48.39 \\
    & Llama-3-70B & 0.00 & 0.00 & 62.50 &  75.00 \\   
    \midrule
    \multirow{4}{*}{High School Physics} & GPT-4-Turbo &6.29 & 31.25 & 6.99 & 30.30 \\
    &  GPT-4o &  13.33 & 23.81 & 23.26 & 38.46 \\
    & Claude-3-Opus & 5.75 & 35.71 & 21.28 & 38.46 \\
    & Llama-3-70B & 31.25 & 26.32 & 18.52 &  27.27 \\
    \midrule
    \multirow{4}{*}{Machine Learning} & GPT-4-Turbo &15.72 & 9.26 & 19.69 & 40.98 \\
    &  GPT-4o & 37.31 & 20.00 & 43.48 & 33.09 \\
    & Claude-3-Opus &  27.03 & 47.62 & 34.88 & 39.06 \\
    & Llama-3-70B & 12.82 & 12.82 & 41.67 & 57.47 \\ 
    \midrule
    \multirow{4}{*}{Professional Law} & GPT-4-Turbo &21.74 & 16.85 & 23.08 & 22.47 \\
    &  GPT-4o & 10.87 & 6.25 & 10.87 & 18.29 \\
    & Claude-3-Opus & 26.12 & 32.97 & 21.74 & 29.41 \\
    & Llama-3-70B & 43.65 & 45.45 & 25.00 &  27.78\\ 
    \midrule
    \multirow{4}{*}{Public Relations} & GPT-4-Turbo &25.97 & 21.28 & 36.23 & 60.00 \\
    &  GPT-4o & 18.87 & 44.44 & 27.03 & 32.61 \\
    & Claude-3-Opus & 20.55 & 28.30 & 35.09 & 49.18 \\
    & Llama-3-70B & 0.00 &   10.64 & 12.50 &  12.20\\ 
    \midrule
    \multirow{4}{*}{Virology} & GPT-4-Turbo &82.97 & 64.39 & 81.63 & 66.29 \\
    &  GPT-4o & 86.67 & 69.03 & 87.80 & 75.37 \\
    & Claude-3-Opus & 84.39 & 76.36 & 83.76 & 79.42 \\
    & Llama-3-70B & 6.85 & 6.49 & 6.25 &  6.49\\
    \bottomrule
  \end{tabular}
  }
\end{table}

\begin{table}[htp]
  \caption{Comparison of Zero-Shot, Chain-of-Thought (CoT), and Few-Shot Recall score on 10 \dataset subjects.}
  \label{gpt-recall-score-table}
  \centering
  \scalebox{0.9}{
  \begin{tabular}{lccccc}
    \toprule
    \textbf{Dataset} & \textbf{Models}&{\bf Zero-Shot} & {\bf Zero-Shot CoT} & {\bf Few-Shot} & {\bf Few-Shot CoT} \\
    \midrule
    \multirow{4}{*}{College Chemistry}  & GPT-4-Turbo & 41.86 & 24.00 & 47.22 & 48.00 \\
     & GPT-4o & 52.94 & 16.00 & 50.00 & 20.00 \\
    & Claude-3-Opus & 40.00 & 48.00 & 50.00 & 24.00 \\
    & Llama-3-70B & 12.00 & 8.00 & 8.00 & 12.00 \\
    \midrule
    \multirow{4}{*}{College Mathematics}  & GPT-4-Turbo &0.00 & 0.00 & 0.00 & 0.00 \\
    &  GPT-4o &0.00 & 0.00 & 0.00 & 0.00 \\
    &Claude-3-Opus &  5.00 & 0.00 & 0.00 & 0.00 \\
    & Llama-3-70B & 0.00 & 0.00 & 0.00 & 0.00 \\
    \midrule
    \multirow{4}{*}{Econometrics} & GPT-4-Turbo &5.00 & 33.33 & 8.57 & 100.00 \\
    &  GPT-4o & 11.11 & 0.00 & 28.57 & 100.00\\
    & Claude-3-Opus &  8.70 & 66.67 & 30.00 & 100.00 \\
    & Llama-3-70B & 0.00 & 0.00 & 0.00 & 100.00 \\
    \midrule
    \multirow{4}{*}{Formal Logic} & GPT-4-Turbo &14.89 & 23.08 & 23.33 & 23.08 \\
    &  GPT-4o & 36.36 & 7.69 & 40.00 & 23.08 \\
    & Claude-3-Opus & 23.53 & 30.77 & 50.00 & 30.77 \\
    & Llama-3-70B & 0.00 & 0.00 & 0.00 & 0.00 \\
    \midrule
    \multirow{4}{*}{Global Facts} & GPT-4-Turbo &15.79 & 25.00 & 23.53 & 33.33 \\
    &  GPT-4o & 16.67 & 8.33 & 28.57 & 16.67 \\
    & Claude-3-Opus & 33.33 & 41.67 & 44.44 & 50.00 \\
    & Llama-3-70B &0.00 & 0.00 & 0.00 & 25.00 \\
    \midrule
    \multirow{4}{*}{High School Physics} & GPT-4-Turbo &5.13 & 66.67 & 5.71 & 66.67 \\
    &  GPT-4o &  11.11 & 33.33 & 20.00 & 66.67 \\
    & Claude-3-Opus & 4.76 & 66.67 & 18.18 & 66.67 \\
    & Llama-3-70B &33.33 & 33.33 & 66.67 & 33.33 \\
    \midrule
    \multirow{4}{*}{Machine Learning} & GPT-4-Turbo &13.51 & 9.09 & 17.24 & 45.45 \\
    &  GPT-4o & 35.71 & 18.18 & 66.67 & 36.36 \\
    & Claude-3-Opus &  24.00 & 54.55 & 37.5 & 45.45 \\
    & Llama-3-70B &0.00 & 9.09 & 0.00 & 9.09 \\
    \midrule
    \multirow{4}{*}{Professional Law} & GPT-4-Turbo &20.00 & 16.67 & 21.43 & 22.22 \\
    &  GPT-4o & 14.29 & 5.56 & 14.29 & 16.67 \\
    & Claude-3-Opus & 24.14 & 33.33 & 28.57 & 27.78 \\
    & Llama-3-70B & 5.56 & 5.56 & 5.56 & 5.56 \\
    \midrule
    \multirow{4}{*}{Public Relations} & GPT-4-Turbo &23.53 & 22.22 & 33.33 & 66.67 \\
    &  GPT-4o & 18.18 & 44.44 & 28.57 & 33.33 \\
    & Claude-3-Opus & 18.75 & 33.33 & 33.33 & 66.67 \\
    & Llama-3-70B &11.11 & 11.11 & 11.11 & 11.11 \\
    \midrule
    \multirow{4}{*}{Virology} & GPT-4-Turbo &88.37 & 59.65 & 85.11 & 61.40 \\
    &  GPT-4o & 92.86 & 64.91 & 97.30 & 71.93 \\
    & Claude-3-Opus & 88.89 & 73.68 & 94.29 & 77.19 \\
    & Llama-3-70B & 38.60 & 40.35 & 36.84 & 52.63 \\
    \bottomrule
  \end{tabular}
  }
\end{table}

\clearpage

\section{Heterogeneity of Errors in MMLU Subsets -- Full List}
\label{sec: Appendix-B}
Here, we provide an extensive list of qualitative observations from the manually validated MMLU subsets. Some of these observations were previously reported in by the AI Explained YouTube Channel~\citep{aiexplained} and a Medium blog post~\citep{medium}.

\begin{description}[leftmargin=0pt,noitemsep,nolistsep]
\item[Professional Law] -- There are several contextual limitations that pose challenges for accurate question answering. One major issue is the lack of specificity regarding jurisdictions. The benchmark does not clearly distinguish between different jurisdictions, despite focusing on U.S. law. Additionally, the inherently interpretative nature of legal principles means that there is often no definitive ground truth, making it difficult to provide unequivocal answers.
\item[Professional accounting] -- These questions mainly come from \textit{FAR CPA Exams}, and are of high quality. There are minor issues in scraping where numeric answers are not converted properly and significance is lost, \eg where the given correct answer is \$242,000, while the correct computed answer is \$241,843. Furthermore, like in professional law, all questions assume U.S. professional accounting practice, even though this is rarely specified. 
\item[Human Aging] -- The questions are mostly correct, except for some questions containing underspecified information. \eg ``In \textit{this chapter's Senior View}, Dr. Shealy advises you to''. 
\item[Global Facts] -- Almost all questions needed consulting external sources to be answered, such as \url{ourworldindata.org} (18 cases) and \url{pewresearch.org} (15 cases); in a few cases multiple sources were providing conflicting answers to the same question -- for example, on the perceived corruption of political parties in 2013 Ethiopia, \url{ourworldindata.org} confirms the answer in MMLU, while other sources such as the Global Corruption Barometer from Transparency International were providing conflicting answers.
\item[Virology] -- Incorrect ground truths labels are particularly prevalent within the Virology subset. Many of the incorrect labels are for relatively simple questions, such as identifying the description of a pandemic, this suggests errors may stem from problems parsing the original datasets (in most cases the \textit{Human Virology} textbook's student resources). In addition, there is a range of issues relating to question and option clarity where the necessary context required to answer the question is missing, such as which family of diseases is being referred to.
\item[Business Ethics] -- This subset includes several unclear questions where respondents must identify multiple correct statements in a listed statement. MMLU provides only the last statement of the list as the question, instead of including the entire question.
\item[Philosophy]  -- Several samples in this subset cannot be explicitly found in an external source. 
\item[Public Relations] -- This subset includes various errors, ranging from multiple correct answers and wrong ground truth to bad question clarity. 
\item[Anatomy] -- This is almost entirely correct. 
\item[Abstract algebra] -- This subset contains several answers that are simply incorrect. Some answers are ambiguous because they assume a ring does not need to have multiplicative identities, which is usually assumed in modern treatments of rings. This changes the outcome of the given answer. 
\item[College Chemistry] -- The questions were sourced from textbooks (\eg \textit{Chechik: Electron Paramagnetic Resonance}, \textit{Hore: Nuclear Magnetic Resonance 2e}) and standardised college-level exams (\eg \textit{GRE Chemistry Practice Test}). We identified erroneous questions resulting from simple parsing errors and unknown causes. For example, questions spanning multiple lines in the original source were often parsed incorrectly, leading to a part of the question being presented as the first MMLU choice (Option A) and the exclusion of Option D. Additionally, some questions originally included option E, which was the correct answer, but this option was omitted from the MMLU choices to fit into the 4-choices question style. Furthermore, there were questions with ground truth labels that did not match the answers provided in the source, with no apparent cause for the discrepancy.
\item[College Medicine] -- The questions were mostly sourced from textbooks (\eg \textit{Maughan \& Gleeson: The Biochemical Basis of Sports Performance 2e}) and standardised college-level medical exam (\eg MCAT). The majority of question-answer pairs are of good quality. However, questions that were sourced from \textit{Maughan \& Gleeson: The Biochemical Basis of Sports Performance 2e} can also be found in the Clinical Knowledge subject.
\item[Clinical Knowledge] -- The questions were mostly sourced from textbooks (\eg \textit{Maughan \& Gleeson: The Biochemical Basis of Sports Performance 2e}, \textit{Cox \& Roper: Clinical Skills}, \textit{Endacott, Jevon \& Cooper: Clinical Nursing Skills Core and Advanced}). The majority of question-answer pairs are of good quality. One specific question was annotated very well by adding the time when the question was asked (\ie ``Which of the following statements is true about informal carers \textbf{(as of 2020)}?'') which correctly indicates the ever-changing nature of these questions. However, questions sourced from \textit{Maughan \& Gleeson: The Biochemical Basis of Sports Performance 2e} can also be found in the College Medicine subject.
\item[Formal Logic] -- The dataset contains a significant number of questions with incorrect answers. These are primarily sourced from the `Oxford Learning Link' website. Inaccuracies are not because of invalid scraping: For example, one question states that $(F\wedge L)\wedge \neg C$ is correct, but $F\wedge L \wedge \neg C$ is not, even though these two formulas are clearly equivalent. 
\item[Logical fallacies] -- Some questions discuss fallacies, \eg the \textbf{solid} slope fallacy,  which come from a set of flashcards obtained from the `Quizlet' website, but otherwise do not return any hits on Google. There is also a large set of unclear questions involving an argument with a logical fallacy, but without any question. For instance, one ``question'' is ``All things that are spoiled are inedible. Timothy is spoiled. So, Timothy is inedible.'' The original source of this question also had the instruction ``Select the fallacy-type which best describes the reasoning contained in each of the passages below.'', but this was lost when adding it to the MMLU dataset.
\item[Machine Learning] -- Most questions were sourced from exam papers, assignments or online quizzes. About 30\% of the questions require expert knowledge and reasoning. The main issues of this subset are bad question clarity and bad options clarity. \eg some quiz questions are based on past knowledge, and the descriptions in the questions may be vague or inapplicable today. 
\item[Electrical Engineering] -- All the questions are sourced from the `Electrical4U' website. However, 2 answers have been incorrectly extracted from the website. 
\item[College Mathematics] -- The majority of the questions are from GRE papers, but they have been modified to have 4 options instead of 5. The majority of the questions require expert knowledge and reasoning. However, there is one question that was incorrectly adjusted when changing from 5 options to 4.
\item[College Computer Science] -- The majority of the questions are from a Computer Science GRE practice exam and online boards with practice questions.
\item[High School Mathematics] -- The majority of the questions are extracted from AP Mathematics and require expert knowledge and reasoning.
\item[High School Statistics] -- The majority of the questions are from AP Statistics and can be directly obtained from the \url{crackap.com} website. However, two answers were incorrectly extracted. 
\item[Miscellaneous] -- The questions are direct facts regarding celebrities, movies, and global knowledge, which can be directly extracted from the internet. However, there were some incorrect answers and vague questions that could be difficult to answer accurately. For instance: `How many cups of coffee were consumed in the United States in the past week (as of 2013)?'
\item[Econometrics] -- The majority of the questions are correct, but some questions contain unverifiable references. (E.g., 'Consider again the VAR model of equation 16,' but equation 16 cannot be found within the question.)
\item[High School Chemistry] -- This is almost completely correct. 
\item[High School Physics] -- This subset is almost entirely correct. The majority of the questions are from AP Statistics and can be directly obtained from the `crackap' website.
\item[College Physics] -- Some questions (approximately 20\%) were duplicated.
\item[High School Geography] -- This is almost completely correct.
\item[Elementary Mathematics] -- This is almost entirely correct. There are only a few questions (mostly basic equations) with wrong ground truth answers.
\item[Conceptual Physics] -- This is almost completely correct, with only a few questions having wrong ground truth answers due to issues with data scraping from unreliable sources.
\item[Astronomy] -- Most questions and their corresponding answers are correct. A unique source was not identified, but some can be found on `Quizlet.' Two questions are unclear because they are missing essential details. Additionally, two question options are confusing due to incorrect parsing of the power of 10 (e.g., $10^{16}$ is incorrectly shown as $1016$). One question lacks a correct answer because it is outdated, highlighting the importance of keeping questions updated with recent discoveries in this field.
\item[Professional Psychology] -- This subset mainly contains questions in the style of Examination for Professional Practice in Psychology (EPPP) mainly used in most U.S. states and Canadian provinces. Among 100 randomly sampled questions, there are four erroneous instances. For instance, one of them contains no correct answer due to incorrect parsing (\ie incorrectly parsing `0.12' as `0.42'). Another example is a bad question clarity where the answer is highly dependent on the geographical context (\ie `In most cases, statutory responsibility for establishing minimal standards for professional competency to protect the public from harm rests with ...').
\item[Security Studies] -- The questions are sourced from textbooks (\eg `\textit{Collins: Contemporary Security Studies 6e Student Resources}'). Six out of 100 instances have bad question clarity due to underspecified abbreviations (\eg `HM' and `TNC'). When read in this form without other supporting context, readers may confuse the abbreviation with other relevant concepts in the security studies. In the textbook, the questions were presented as part of a specific chapter, and the abbreviations were defined in the earlier questions.
\item[High School Computer Science] -- This subset is completely clean.
\item[High School Psychology] -- The questions are mostly sourced from AP Psychology practice questions. Among the 100 instances, there are four problematic ones. Three of them are due to missing statements. These questions are presented as multiple true-false questions (\ie the options are `I only', `II only', `III only', and `I and II only') without the four statements. One of the four erroneous instances is problematic due to racial stereotyping (\ie `\textit{Aisha is a beautiful, black teenager. If she is \textbf{typical}, she most likely believes that}').
\item[Moral Disputes] -- This subset contains four erroneous instances. For instance, three of them refers to underspecified entities (\eg `\textit{\textbf{Ashford's article} is meant to address a particular paralysis in the face of}', `\textit{If \textbf{Thomson's conclusion} is correct, then}', `\textit{In responding to terrorism, says \textbf{Walzer}, it is particularly important that}').
\item[Moral Scenarios] -- This subset contains three instances with ambiguous questions requiring more background information before deciding whether the statement is `morally wrong'.
\item[High School World History] -- The questions are sourced from publicly available AP World History practice questions from sources such as `crackap'. This subset is almost completely correct, except for one question with a wrong ground truth which may be caused by a random mistake.
\item[High School European History] -- The errors in this subset are mainly due to bad option clarity. In some cases, there is an option that is equivalent to the correct one. Additionally, in a few samples, the passage provided as a reference does not contain the answer to the question, despite the question implying that it should.
\item[Computer Security] -- The questions are sourced from university-level quizzes, with most of them being correct. There are three instances with wrong ground truth answer, however it is unclear what may have caused it.
\item[Prehistory] -- Completely clean.
\item[US Foreign Policy] -- The questions were mostly sourced from textbooks (\eg \textit{Cox \& Stokes: US Foreign Policy 3e}) and publicly available final exams. There is only one question with bad clarity in terms of the temporal context (\ie `How many states in the international system are likely to have nuclear weapons \textbf{right now}?').
\item[Sociology] -- This subset contains questions that are sourced from textbooks (\eg `\textit{Fulcher \& Scott: Sociology 4e}'). There are two instances with bad question clarity, mainly because the question requires time and geographical contexts (\ie `\textit{Patterns of drug use in Britain reveal that:}' and `\textit{Which of the following policies did the New Labour government not pursue?}').
\item[World Religions] -- This subset is almost completely correct. One instance with a bad question clarity requires more context about when is this question asked (\ie `\textit{When was the \textbf{current} Dalai Lama born?}').
\item[College Biology] -- This subset is mostly sourced from a publicly available GRE Biology practice test. There are two erroneous instances; one of them has a wrong ground truth answer, while the other one has its ground truth answer copy pasted to the question due to incorrect parsing (\ie `\textit{Stabilization of the unique coiled structure of an alpha helix in a protein is primarily attributed to \textbf{(A) hydrogen bonding between the peptide backbone atoms}}').
\item[High School Biology] -- Most questions in this subset are sourced from AP Biology practice questions. There are five erroneous instances, three of them seem to be caused by bad parsing. For instance, the correct option A is simply written as `0'. Another erroneous example is where the question refers to a previous question (\ie `\textit{\textbf{Refer to question 15 for details on the squirrel population.} Which of the following conditions is required to keep this population in Hardy-Weinberg equilibrium?}').
\item[International Law] -- Most questions in this subset are sourced from a textbook (\ie `Bantekas \& Papastavridis: International Law Concentrate 4e'), with most of them are correct.
\item[Jurisprudence] -- Most questions in this subset are sourced from a textbook (\ie `Wacks: Understanding Jurisprudence 5e'), with most of them are correct.
\item[Medical Genetics] -- This subset is sourced from university-level final exams and is completely correct.
\item[Professional Medicine] -- This subset contains questions that are sourced from the publicly available sample questions of a standardised medical exam in the US (\ie the United States Medical Licensing Examination (USMLE)). Several questions with an unknown source can be answered with several searches on established websites (\eg National Center for Biotechnology Information (NCBI), Mayo Clinic). This subset only contains one erroneous instance due to bad question clarity where the original question has an accompanying photograph. 
\item[Marketing] -- The questions in this subset are sourced from university-level quizzes and textbooks. There are four erroneous instances. One question misses the temporal context, while the other is resented as a statement with unclear relation with the options.
\item[High School Microeconomics] -- The questions in this subset are sourced from AP Microeconomics. Most of them are correct, except for two instances. One instance has a wrong ground truth answer for unknown reason. Another one has a bad question clarity, referring to unspecified question (\ie `\textit{The basis for the answer in \textbf{number 1} is}').
\item[Management] -- This is almost completely correct.
\item[Nutrition] -- This is almost entirely correct. Only one question has a minor mistake in one of its options (\ie the `>' sign was mistakenly replaced by `?' mark) and one question was annotated by adding the time when the question was asked (similar to Clinical Knowledge) to indicate the ever-changing nature of the questions.
\item[Human Sexuality] -- Most questions in this subset are sourced from university-level exams that are publicly available. There are 17 erroneous instances of various types. Several questions are highly subjective or speculative (\eg `\textit{Persons with liberal attitudes about premarital sex are likely to:}', `\textit{Alexander the Great was probably}' (in relation to his sexuality)).
\item[High School Macroeconomics] -- The questions in this subset are sourced from AP Macroeconomics. Most of them are correct, except for three instances. One instance has two identical correct answers. The other two are fill-in-the-blank type of questions, however the position of the blanks are not specified.
\item[High School Government and Politics] -- The questions are mostly sourced from AP U.S. Government and Politics practice questions, with most answers can be found on the `crackap' website. This subset is completely correct.
\item[High School US History] -- The questions in this subset are sourced from AP U.S. History practice questions, with all answers can be found on the `crackap' website. This subset is completely correct.

\end{description}

\section{Error Detection via Fine-tuning}
\label{sec: Appendix-E}
To validate our fine-tuning strategy for error detection, we developed LabelChaos, a dataset designed to mirror the error distribution of the original MMLU. This dataset serves as a benchmark for finetuning models, which are subsequently evaluated on \dataset.

To create LabelChaos, we selected and merged six manually labelled datasets. We chose datasets annotated by humans  (\citep{arc_dataset, OpenBookQA_dataset, mathqa_dataset, medqa_dataset, truthfulqa_dataset}) to avoid the potential inaccuracies associated with automated labelling procedures. After standardising these datasets to align with the format of MMLU, we generate a corrupted version of each by introducing specific types of corruption, as categorised in our annotation protocol (Fig. \ref{fig:mmlu-redux-stats}). These corruptions were carefully applied to replicate the quality and distribution characteristics of the MMLU dataset. The final dataset comprises approximately 264,000 samples, similar to those in MMLU. Below, we provide an overview of the LabelChaos subsets and the corruption procedures used for each type of error.

\begin{itemize}[leftmargin=*]
\item \textbf{Wrong Ground Truth}: The correct answer label is replaced by randomly selecting from the incorrect answers.
\item \textbf{Poor Question Clarity}: An LLM-based corruption is introduced by prompting GPT-3.5 Turbo to modify the question, increasing its difficulty. Few-shot examples from MMLU illustrating the ``poor question clarity'' error type are used.
\item \textbf{No Correct Answers}: The correct answer is replaced with `all options listed' or `all of the above'.
\item \textbf{Unclear Options}: Since most instances of unclear options in the original MMLU stem from parsing errors, we simulate these by introducing parsing errors into the choices, such as splitting strings at incorrect characters.
\item \textbf{Multiple Correct Answers}: GPT-3.5 Turbo is used to generate a new option semantically identical to the correct one. One of the incorrect options is then replaced with this newly generated option. 
\item \textbf{Correct}: The original, uncorrupted dataset.
\end{itemize}

We fine-tune the Llama-3 (8B-Instruct)~\citep{llama3modelcard} using LabelChaos datasets. To balance the distribution in \dataset, where most instances are labelled as "\textit{correct}", we adjusted the label distribution to: 0.1 (Wrong Ground Truth), 0.1 (Poor Question Clarity), 0.1 (No Correct Answers), 0.1 (Unclear Options), 0.1 (Multiple Correct Answers), and 0.5 (correct). For consistency, we used the previously described CoT Prompt as input, with outputs labelled as either ``\textbf{ok}'' or ``\textbf{not ok}''. The training involves 2048 steps, with a batch size of 64, utilising the AdamW optimizer~\citep{loshchilov2018decoupled} with a learning rate of $2\times10^{-4}$ and no weight decay. Due to computational constraints, we apply LoRA~\citep{hu2021lora}, with a rank of 16, to all models. All experiments were conducted on a single Nvidia A100 (40GB) GPU. We present the results of the fine-tuning method in Table \ref{tab:fine-tuning-labelchaos}.

\begin{table}[htp]
    \caption{Comparison of Different Methods and Models for Error Detection in \dataset Scores.}
  \label{tab:fine-tuning-labelchaos}
  \centering
  \begin{tabular}{rccc}
    \toprule
   \textbf{Methods} & \textbf{Recall} & \textbf{F1 Score} & \textbf{F2 Score} \\
    \midrule
   GPT-4 Turbo (Few-shot CoT) &  46.68 & 31.60 & 36.58 \\
    GPT-4 Turbo (Few-shot CoT) &  38.47 & 29.26 & 31.36 \\
     Claude 3 Opus (Few-shot CoT)&  48.85 & 24.03 & 40.29 \\
     Llama3-70B (Few-shot CoT)&  24.87 & 23.16 & 23.10 \\
     Llama3-8B (Fine-tuning)&  56.58 & 34.06 & 44.75 \\
    \bottomrule
  \end{tabular}
\end{table}

\end{document}